\documentclass[sigconf]{acmart}

\usepackage{amsmath,amsfonts} 
\usepackage{colortbl}
\usepackage{xcolor}
\usepackage{tikz}
\usepackage[inline,shortlabels]{enumitem}
\usepackage{xspace}
\usepackage{multirow}

\usepackage[ruled,linesnumbered]{algorithm2e}
\usepackage{inputenc}
\usepackage{mathrsfs}
\usepackage{pgfplots}
\usepackage{tcolorbox}
\usepackage{graphicx}
\usepackage{lettrine}
\usepackage{pifont}
\usepackage{tabularx}
\usepackage{makecell}
\usepackage{booktabs}
\usepackage{diagbox}
\usepackage{float}

\usepackage{array}
\newcolumntype{Y}{>{\centering\arraybackslash}X}

\graphicspath{{figures/}}

\newcommand{\progressbar}[2][0.05]{%
  \begin{tikzpicture}[baseline=-0.5ex]
    \pgfmathsetlengthmacro{\W}{#1\columnwidth}
    \pgfmathsetlengthmacro{\H}{1ex}
    \draw[line width=0.25pt, rounded corners=0.6pt] (0,0) rectangle (\W,\H);
    \fill (0,0) rectangle ({#2*\W},\H);
  \end{tikzpicture}%
}
\newcommand{\pbEmpty}{\progressbar{0}}
\newcommand{\pbHalf}{\progressbar{0.5}}
\newcommand{\pbFull}{\progressbar{1}}

\newcolumntype{P}[1]{>{\centering\arraybackslash}p{#1}}

\definecolor{ccr}{RGB}{120,29,125}  

\hypersetup{
    hypertex=true,
    colorlinks=true,
    linkcolor=ccr,
    anchorcolor=ccr,
    citecolor=ccr,
    filecolor=ccr,
    urlcolor=ccr
}

\def\ie{\textit{i.e.},~}

\def\eg{\textit{e.g.},~}
\def\etal{et al.~}

\def\namerm{OpsAgent}
\def\name{\textit{\namerm}\xspace}

\newlist{questions}{enumerate}{1}
\setlist[questions]{wide=0pt, label=\textbf{RQ\arabic*.}, ref=RQ\arabic*}

\copyrightyear{2026}
\acmYear{2026}
\setcopyright{none}             
\renewcommand\footnotetextcopyrightpermission[1]{} 
\acmConference[ASE '26]{41st IEEE/ACM International Conference on Automated Software Engineering}{October 12--16, 2026}{Munich, Germany}
\acmBooktitle{Companion Proceedings of the 41st IEEE/ACM International Conference on Automated Software Engineering (ASE '26), October 12--16, 2026, Munich, Germany}

\begin{document}

\title{OpsAgent: An Evolving Multi-agent System for Incident Management in Microservices}


\author{Yu Luo}
\affiliation{%
  \institution{Nankai University}
  \city{Tianjin}
  \country{China}
}

\author{Jiamin Jiang}
\affiliation{%
  \institution{Nankai University}
  \city{Tianjin}
  \country{China}
}

\author{Jingfei Feng}
\affiliation{%
  \institution{Nankai University}
  \city{Tianjin}
  \country{China}
}

\author{Lei Tao}
\affiliation{%
  \institution{Nankai University}
  \city{Tianjin}
  \country{China}
}

\author{Qingliang Zhang}
\affiliation{%
  \institution{Nankai University}
  \city{Tianjin}
  \country{China}
}

\author{Xidao Wen}
\affiliation{%
  \institution{Alibaba Cloud}
  \city{Beijing}
  \country{China}
}

\author{Yongqian Sun}
\authornote{Yongqian Sun is the Corresponding author.}
\affiliation{%
  \institution{Nankai University}
  \city{Tianjin}
  \country{China}
}

\author{Shenglin Zhang}
\affiliation{%
  \institution{Nankai University}
  \city{Tianjin}
  \country{China}
}

\author{Tong Liu}
\affiliation{%
  \institution{Lenovo}
  \city{Tianjin}
  \country{China}
}

\author{Wenjie Zhang}
\affiliation{%
  \institution{Lenovo}
  \city{Tianjin}
  \country{China}
}

\author{Dan Pei}
\affiliation{%
  \institution{Tsinghua University}
  \city{Beijing}
  \country{China}
}

\renewcommand{\shortauthors}{Luo, et al.}

\begin{abstract}
  Incident management (IM) is central to the reliability of large-scale microservice systems. 
Yet manual IM, where on-call engineers examine metrics, logs, and traces is labor-intensive and error-prone in the face of massive and heterogeneous observability data. 
Existing automated IM approaches often struggle to generalize across systems, provide limited interpretability, and incur high deployment costs, which hinders adoption in practice.
In this paper, we present \name, a lightweight, self-evolving multi-agent system for IM that employs a training-free data processor to convert heterogeneous observability data into structured textual descriptions, along with a multi-agent collaboration framework that makes diagnostic inference transparent and auditable.
To support continual capability growth, \name also introduces a dual self-evolution mechanism that integrates internal model updates with external experience accumulation, thereby closing the deployment loop.
Comprehensive experiments on the OPENRCA \cite{openrca} benchmark demonstrate state-of-the-art performance and show that \name is generalizable, interpretable, cost-efficient, and self-evolving, making it a practically deployable and sustainable solution for long-term operation in real-world microservice systems.
Notably, its deployment in Lenovo's production environment further validates its effectiveness in real-world industrial settings.
\end{abstract}



\maketitle


\section{Introduction}\label{intro}

Microservice systems have become the de facto architecture for building modern software services, with wide deployments across industries such as IT, government, and finance \cite{rcacopilot}. 
However, incidents (\eg service disruptions and outages) \cite{art,rcacopilot} are inevitable due to the complexity of microservice systems, often resulting in catastrophic economic and operational consequences.
For instance, on June 12, 2025, a faulty quota-control deployment in Google Cloud triggered a global outage that lasted nearly eight hours, disrupting more than 80 GCP services and cascading into failures across e-commerce, finance, AI applications, entertainment platforms, and transportation systems worldwide. 
The economic impact of this incident was substantial, as it encompassed not only Google's direct losses but also widespread hidden costs borne by countless enterprises and end users affected by the disruption \cite{google_crash}.
Thus, comprehensive \underline{i}ncident \underline{m}anagement (IM) that integrates \underline{a}nomaly \underline{d}etection (AD), \underline{f}ailure \underline{t}riage (FT), and \underline{r}oot \underline{c}ause \underline{l}ocalization (RCL) is essential to recover from such disruptions \cite{rca_survey, rcacopilot}.
Traditionally, on-call engineers (OCEs) manually inspect metrics, logs, and traces to identify the root cause when incidents occur \cite{rcacopilot}. 
This process yields interpretable results, as OCEs reason step by step and accumulate expertise, but it becomes infeasible at scale due to the overwhelming volume and heterogeneity of observability data \cite{automap, art}.

To alleviate this burden, automated IM with AI techniques has been extensively explored, falling into two main categories: deep learning (DL)-based IM \cite{microcause, eadro, medicine, mulan, nezha, diagfusion, art} and large language model (LLM)-based IM \cite{rca_survey, ahmed, icl_rca, comet, las_rca, rcacopilot, mabc}. 
Deep learning-based approaches leverage neural networks trained via supervised \cite{diagfusion, eadro} and self-supervised learning \cite{art, deephunt} to extract complex failure patterns from observability data. 
However, the inherent ``black-box'' nature of neural networks yields predictions without transparent reasoning chains, making them hard for OCEs to trust and adopt. 
Furthermore, these models generalize poorly as they are trained on system-specific data, which usually require costly data collection and retraining when moved to new systems.
LLM-based approaches exploit the strong reasoning and natural-language understanding of LLMs, generating diagnoses directly from incident titles and summaries or by matching to similar historical cases \cite{icl_rca,rcacopilot,ahmed}. 
However, two limitations hinder practical deployment: first, many methods depend on large closed-source models (\eg GPT-4), which impose high deployment costs as well as privacy-exposure risks \cite{icl_rca,rcacopilot}; second, without step-wise inference over raw observability data, the pipeline skews toward shallow similarity matching, which limits accuracy and offers no mechanism to accumulate expertise through continued usage \cite{icl_rca,rcacopilot,ahmed}.
As a result, despite promising research progress and good performance on public datasets, automated IM techniques have yet to achieve widespread adoption in practice, and many companies still rely heavily on manual IM.

Thus, OCEs urgently call for \textbf{\textit{a deployable and sustainable IM approach}}.
Recent advances in multi-agent systems (MAS) suggest a promising direction, as they excel at decomposing complex reasoning tasks through collaboration and have already shown success in software engineering domains such as code generation \cite{mapcoder,codes,paircoder}, quality assurance \cite{axnav,gptlens}, and requirements analysis \cite{mare,elicitron}.
Nevertheless, realizing a practically deployable and sustainable MAS-based IM requires overcoming three key technical challenges.

\textit{\textbf{(C1) How to generalize MAS-based IM under heterogeneous and shifting microservice systems?}}  
Microservice systems vary widely in architecture and observability instrumentation, yielding voluminous and heterogeneous observability data.  
A common practice in existing MAS-based IM approaches is to let specialized agents directly operate on raw observability data \cite{mabc}.
While this design simplifies data ingestion, it forces each agent to rely on different modality-specific inputs (\eg metrics, logs, traces), which can lead them to form divergent understandings of the same incident and make their conclusions difficult to reconcile.
The challenge is to design MAS-compatible data processing pipelines that can transform heterogeneous observability data into coherent representations. 
This is non-trivial, as formats, semantics, and granularities differ drastically across modalities, and naive abstraction may discard diagnostic cues essential for accurate reasoning.

\textit{\textbf{(C2) How to ensure interpretable reasoning in MAS-based IM?}}  
In practice, OCEs cannot rely on predictions alone—they must audit the reasoning process before acting on diagnostic results, since misdiagnosis may lead to cascading incidents.
In a MAS, interpretability hinges on clear role specification and a well-structured collaboration workflow that exposes intermediate steps, where the central challenge is balancing granularity with clarity.
Roles that are too coarse obscure responsibility, whereas overly fine-grained roles impose excessive coordination overhead.
At the same time, poorly designed interaction workflows may lead to ad hoc communications between agents, reducing transparency and hindering human auditing.

\textit{\textbf{(C3) How to enable continual capability growth in MAS-based IM?} } 
Incidents in real-world microservice systems are diverse, evolving with frequent software updates.
Traditional supervised or self-supervised training paradigms produce static models that tend to memorize patterns rather than improve intrinsic diagnostic skills.
Existing MAS-based approaches, such as D-Bot \cite{dbot} and Flow-of-Action \cite{flowofaction}, are built on closed-source LLMs with fixed capabilities (\eg GPT-4) and static SOPs, neither of which provides mechanisms for autonomous learning or capability evolution, making them ill-suited to adapt to novel incidents.
The challenge is thus to enable continual capability growth in MAS-based IM. 
However, enhancing diagnostic capability is difficult because internal parameter updates are often unstable and fail to consolidate acquired experience, while relying solely on external knowledge (\eg SOPs and knowledge base) limits the improvement of the agent's intrinsic reasoning capability.

\begin{figure}[htbp]
    \centering
    \includegraphics[width=\linewidth]{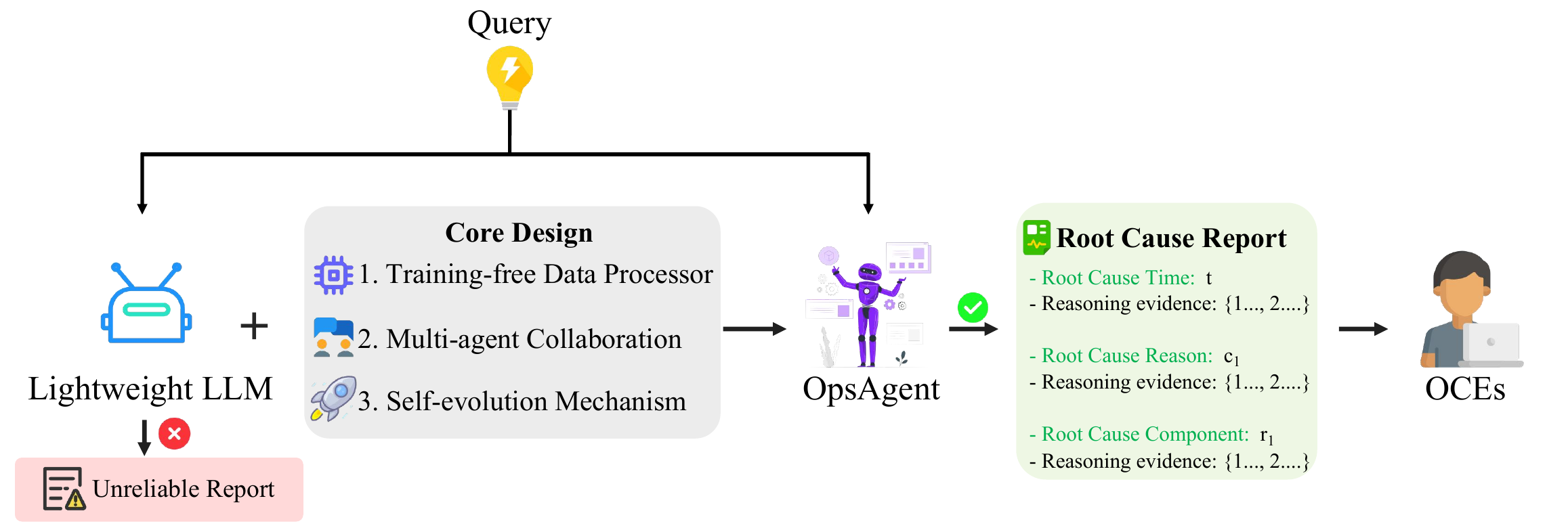}
    \caption{
    \textbf{From lightweight LLM to MAS-based IM}. \name turns a lightweight LLM into a deployable and sustainable IM system by incorporating (1) training-free data processor (Section~\ref{sec:data_processor}), (2) multi-agent collaboration (Section~\ref{sec:multi_agent}), and (3) self-evolution mechanism (Section~\ref{sec:self-evolution}).
    }
    \label{fig:teaser}
\end{figure}

To address these challenges, we present \name (as shown in Fig.~\ref{fig:teaser}), a lightweight and modular MAS for IM.
Instead of relying on massive closed-source models, \name employs a relatively small 14B-parameter model as its reasoning core, and is explicitly designed to support cross-system generalization, interpretable diagnostics, and continual capability growth.
Specifically, \name tackles \textbf{\textit{(C1)}} by introducing a training-free data processor that unifies heterogeneous observability data into structured textual representations, enabling consistent reasoning across diverse environments.
For \textbf{\textit{(C2)}}, it employs a multi-agent collaboration framework that mirrors human problem solving, where specialized roles and coordinated workflows make diagnostic reasoning both transparent and auditable.
For \textbf{\textit{(C3)}}, it integrates a dual self-evolution mechanism that combines reinforcement learning with reflection-based knowledge distillation, allowing the system to progressively enhance its diagnostic capability rather than remain static.
Our contributions are summarized as follows:
\begin{enumerate}[leftmargin=1.5em]
\item We propose \name, a novel MAS-based IM method that is generalizable, interpretable, cost-efficient and self-evolving, which offers a practically deployable and sustainable solution for real-world microservice systems.
\item We introduce a dual self-evolution mechanism that integrates internal model updates with external experience accumulation. This closes the deployment loop by turning validated outcomes into consolidated knowledge and guiding subsequent cases, enabling sustained and auditable capability growth for long-term deployment in dynamic microservice systems.
\item We evaluate \name on the OPENRCA \cite{openrca} benchmark and demonstrate state-of-the-art performance.
Through comprehensive experiments, we also prove its generalizability, interpretability, cost-efficiency and capability of self-evolving.
We also deploy it in Lenovo's production environments to validate its practical utility.
To ensure reproducibility, we release code, prompts and data~\footnote{\url{https://anonymous.4open.science/r/OpsAgent-CCC0}}.
\end{enumerate}

\section{Background}\label{background}
\subsection{Observability in Microservice System}
Microservices architecture is a software system design paradigm where applications are decomposed into a set of loosely coupled, independently deployable services \cite{microservices}. 
This approach improves scalability, agility, and maintainability by allowing each service to evolve and operate independently.
In microservices environments, effective monitoring and understanding of system behavior rely heavily on three fundamental types of observability data:

\begin{itemize}
\item 
\textbf{Metrics} are structured time-series data that quantify system performance and resource usage, which offer a high-level overview of system health and trends.
\item 
\textbf{Logs} are semi-structured text data that record system event details, such as service start, service shut down, and error stack traces.

\item 
\textbf{Traces} are topological records generated by distributed tracing systems that capture the full invocation path of requests across services.
Traces also reveal the sequence, duration, and dependencies of service calls, enabling analysis of cross-service interactions.
\end{itemize}

\subsection{Reinforcement Learning}
Reinforcement Learning (RL) is a machine learning paradigm where an agent learns an optimal policy via repeated environment interactions to maximize cumulative reward.
Among various RL algorithms, \textbf{proximal policy optimization (PPO)} stands out as it is a policy-gradient RL algorithm that improves training stability and sample efficiency by using a clipped objective to limit policy updates, preventing collapse and balancing exploration and exploitation \cite{ppo,gpt4}.


\subsection{Retrieval-Augmented Generation}
While LLMs excel at text generation and semantic comprehension, they have two key limitations \cite{rag_survey}: static knowledge and insufficient specialized domain expertise.
To address these shortcomings, the \textbf{Retrieval-Augmented Generation (RAG)} \cite{graphrag,memorag,hybridrag} framework integrates external knowledge into generation, enabling content production based on retrieval results. 
It retains LLMs' language generation capabilities while enhancing output accuracy and domain adaptability via external knowledge injection, serving as a key technology to boost large models' practicality in professional scenarios.


\subsection{Problem Definition}\label{sec:problem_definition}
In our setting, IM is framed as a natural-language-driven multi-task problem. 
The input consists of a natural language query $q$, which may describe one or more incidents within a given time window and specify a combination of three subtasks: AD, FT, and RCL. 
Let the multimodal observability data be denoted as $\mathcal{X}=(X^M, X^L, X^T)$, where $X^M$, $X^L$, and $X^T$ correspond to metrics, logs, and traces, respectively. 

The AD task aims to identify the root cause occurrence time $t$, \ie the earliest timestamp at which anomalous behavior begins to manifest in the system. 
The FT task involves selecting the most probable failure type $c$ from a predefined category space $\mathbb{C}=\{c_1, c_2, \dots, c_n\}$. 
Finally, the RCL task localizes the most likely faulty component $r$ from the set of root cause components $\mathbb{R}=\{r_1, r_2, \dots, r_m\}$, encompassing entities across different system levels (\eg nodes, services, pods).
The overall objective is to learn a mapping function $ \mathcal{F}:(q, \mathcal{X}) \to (t, c, r)$.

\section{Methodology}
\subsection{Overview}


\name consists of three modules: (1) training-free data processor (Section~\ref{sec:data_processor}), (2) multi-agent collaboration (Section~\ref{sec:multi_agent}), and (3) self-evolution mechanism (Section~\ref{sec:self-evolution}). 
Each module directly targets one of the challenges in Section~\ref{intro}. 
\textbf{\textit{(C1)}} To generalize across heterogeneous deployments, the training-free data processor extracts abnormal patterns from raw metrics, logs, and traces and converts them into unified, semantically aligned textual descriptions that all agents can consume consistently. 
\textbf{\textit{(C2)}} To ensure interpretable reasoning, the multi-agent collaboration framework mirrors human problem solving by defining well-scoped expert roles (\ie AD, FT, and RCL), coordinating them with an orchestrator, and supporting iterative cross-review that produces explicit intermediate evidence and an auditable diagnostic trail. 
\textbf{\textit{(C3)}} To enable continual capability growth, the self-evolution mechanism integrates intrinsic updates via PPO fine-tuning with explicit experience accumulation through agent reflection, which allows \name to improve over time. 
Finally, \name is deliberately lightweight as it operates with a modest 14B-parameter open-source model as the reasoning core, avoiding massive closed-source LLMs and keeping deployment costs low. 
This design supports a practically deployable and sustainable solution for real-world microservice systems.


\subsection{Training-free Data Processor}\label{sec:data_processor}

\begin{figure}[htbp]
    \centering
    \includegraphics[width=\linewidth]{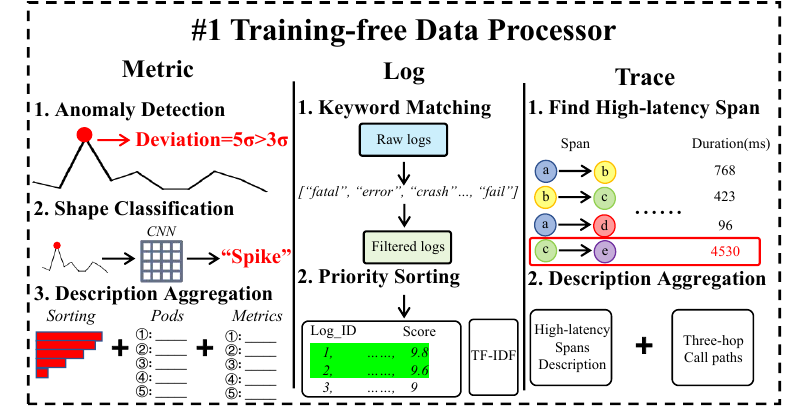}
    \caption{
    \textbf{Training-free Data Processor}. The processor handles three types of observability data separately: metrics (left), logs (middle), and traces (right).
    }
    \label{fig:data_processor}
\end{figure}

\begin{figure}[htbp]
    \centering
    \includegraphics[width=\linewidth]{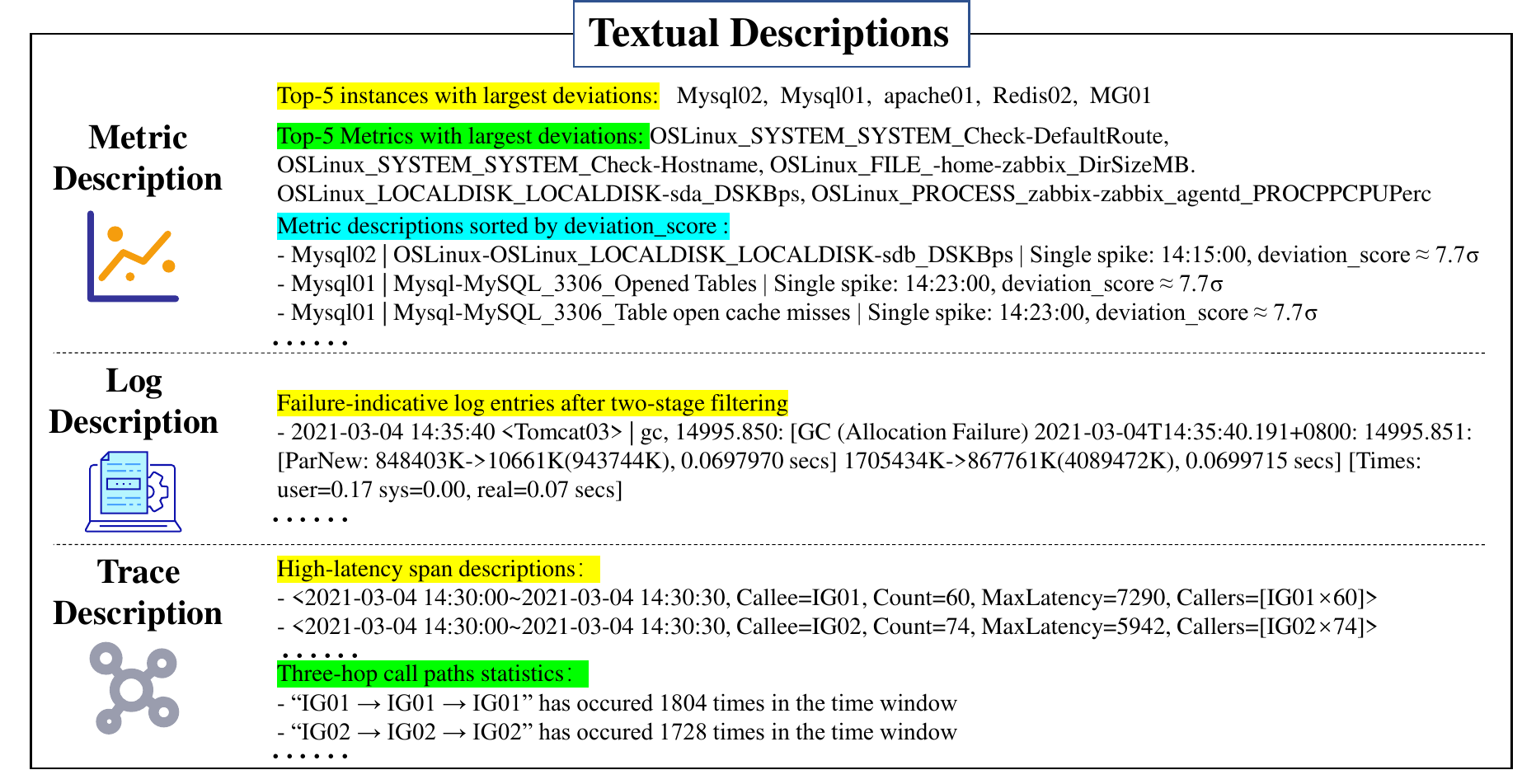}
    \caption{
    Illustrative example of data descriptions.
    }
    \label{fig:data_descriptions}
\end{figure}

Unlike DL-based IM methods that require large-scale data to learn feature distributions \cite{diagfusion, deephunt, art, eadro}, our data processor adopts a training-free approach as shown in Fig.~\ref{fig:data_processor}. 
This design eliminates the need for costly data collection and retraining, while ensuring better generalization across heterogeneous microservice systems. 
The key idea is to extract abnormal patterns from raw metrics, logs, and traces via statistical and heuristic techniques, then convert them into unified textual descriptions that serve as the input for the downstream agents in AD, FT, and RCL.
We presented an illustrative example of the data descriptions in Fig.~\ref{fig:data_descriptions}.

\begin{figure*}[htbp]
    \centering
    \includegraphics[width=0.85\linewidth]{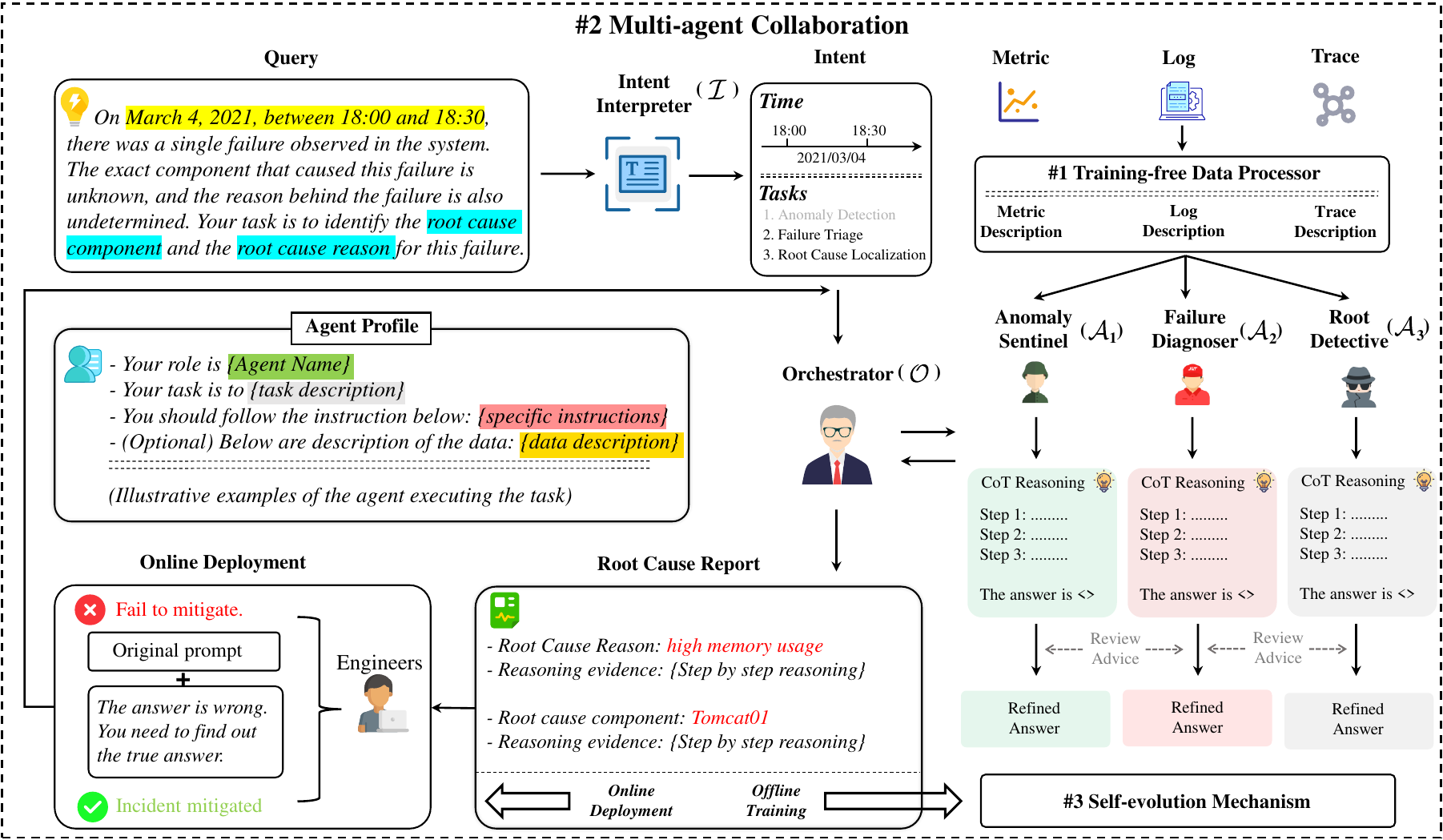}
    \caption{
    \textbf{Multi-agent Collaboration}. Agents with predefined roles (via agent profile) cooperate under a structured \underline{\textbf{workflow}} and \underline{\textbf{cross-review mechanism}} to enhance reasoning from multiple perspectives. The Root Cause Report not only guides \underline{\textbf{online incident mitigation}} but also feeds \underline{\textbf{offline training}}, closing the loop for sustainable capability growth.
    }
    \label{fig:multi-agents}
\end{figure*}

\textbf{Metrics.}  
We process metrics in three stages to filter noise and highlight diagnostically useful signals.  
\textit{(1) Anomaly Detection.} 
Raw metrics are often noisy and periodic, so we first identify statistically significant deviations to reduce volume and concentrate on incident-indicative behaviors. 
We apply the 3-sigma rule within a sliding window, marking samples with 
$\mathrm{score}(x_t)=\frac{|x_t-\mu|}{\sigma}>3$
as anomalies, and retain their deviation scores in units of $\sigma$.  
\textit{(2) Shape Classification.} 
Since different anomaly patterns (\eg spikes, steady increases, level shifts) imply different physical meanings, we use a pre-trained CNN \footnote{We trained this CNN to classify the anomaly shapes, with its design corresponding to the PatternMatcher method \cite{patternmatcher}.} with a context window of 20 steps before and 10 after to assign each anomaly a discrete shape label \cite{patternmatcher}.  
\textit{(3) Description Aggregation.} 
To provide agents with compact yet informative inputs, anomalies are transformed into textual descriptions of the form \textit{$\langle$service\_instance, metric\_name, anomaly\_pattern: timestamp, deviation\_score = $k\sigma \rangle$}. 
We then sort anomalies by deviation score, aggregate scores per service instance to select the top-5 pods, and per metric name to select the top-5 metrics. 
This preserves rich contextual and statistical information while filtering for the most diagnostically relevant signals.

\textbf{Logs.}
We adopt a two-stage pipeline to extract incident-indicative logs while discarding large volumes of irrelevant entries.  
\textit{(1) Keyword Matching.}
Operational logs dominate raw log data, so we first filter by a predefined incident-related lexicon (\eg \textit{fatal}, \textit{error}, \textit{crash}, \textit{fail}) to remove routine operational messages.  
\textit{(2) Priority Sorting.}
Even after keyword filtering, many entries remain and are bursty. 
We parse logs into templates with \textsc{Drain3} \cite{drain} to normalize variable content, then rank templates by TF–IDF \cite{tfidf} so that entries that are salient within the current window yet uncommon are prioritized. 
We keep entries whose templates rank above an adaptive threshold (default: 80th percentile) and deduplicate those sharing the same template within a one-minute window by retaining the earliest instance.  
The resulting log descriptions are the filtered raw entries, preserving full context for downstream agents while emphasizing high-signal, non-redundant evidence.

\textbf{Traces.}
We process traces in two steps to surface latency anomalies and their structural context.
\textit{(1) Find High-latency Span.}
We classify spans as high latency if their latency exceeds a threshold specific to each call type.
Because latency distributions vary across call types, a single global threshold can mislabel normal calls in some services and miss true outliers in others.
We therefore use a flexible per-call-type threshold (default: 95th percentile \cite{trioxpert}) and mark spans above it as high latency.
\textit{(2) Description Aggregation.}
First, to capture co-occurring hotspots and context, we group high-latency spans by 60\,s windows and callee, computing per window the total calls, maximum latency, and caller distribution. 
This yields textual records \textit{$\langle$time\_interval, callee, count, max\_latency, callers$\rangle$}, which highlight when and where latency concentrates.
Second, to reveal recurrent patterns that may indicate systemic bottlenecks, we extract three-hop call paths (\textit{grandparent $\to$ caller $\to$ callee}) from high-latency spans and tally their global frequencies.
Together, these descriptions preserve essential temporal and topological cues while suppressing routine or low-latency traffic, providing focused representation for downstream agents.

\subsection{Multi-agent Collaboration}\label{sec:multi_agent}
As illustrated in Fig.~\ref{fig:multi-agents}, this module has three parts. 
We first outline the workflow from a natural-language \textit{Query} to a \textit{Root Cause Report} coordinated by the \textit{Orchestrator} and expert agents. 
We then describe the cross-review mechanism that lets agents critique and refine one another's reasoning to improve accuracy and auditability. 
Finally, we explain how online deployment and offline training use the \textit{Root Cause Report} to close the loop and sustain capability growth.

\subsubsection{Workflow}
All agents in \name are defined through an \textit{agent profile}, a structured prompt template that specifies the agent's name, task description, operational instructions, and illustrative examples. 
This unified specification ensures consistent behavior and sets the stage for the workflow below:
Given a \textit{Query}, the \textit{Intent Interpreter} ($\mathcal{I}$) extracts the analysis time window and the requested tasks (AD, FT, RCL).
The \textit{Orchestrator} ($\mathcal{O}$) then retrieves metrics, logs, and traces within that window and invokes the training-free data processor to convert them into unified, semantically aligned descriptions. 
This normalization provides a shared evidence base for all agents, improving consistency and auditability.
With roles aligned to diagnostic tasks rather than data modalities, three expert agents—\textit{Anomaly Sentinel} ($\mathcal{A}_1$) for AD, \textit{Failure Diagnoser} ($\mathcal{A}_2$) for FT, and \textit{Root Detective} ($\mathcal{A}_3$) for RCL—reason over the same descriptions using Chain-of-Thought prompting \cite{cot} to produce task-specific answers together with their stepwise rationales in parallel. 
Mapping roles to AD/FT/RCL avoids early information asymmetry (\eg splitting by metrics/logs/traces would deprive each agent of complementary signals) and mirrors real operational practice \cite{srebook}, thereby supporting interpretable collaboration. 
Then, the \textit{Orchestrator} coordinates cross-review to reconcile disagreements and surface missing evidence (details in the next subsection). 
When refinement terminates, it compiles a \textit{Root Cause Report} that records the final results and the intermediate reasoning evidence, enabling human auditing and directly addressing (\textbf{\textit{C2}}).

\subsubsection{Cross-review Mechanism}
Cross-review leverages the fact that the three expert agents—$\mathcal{A}_1$, $\mathcal{A}_2$, and $\mathcal{A}_3$—bring distinct yet complementary perspectives to the same incident. 
A single perspective can miss context or overfit local evidence, whereas peer critique helps surface gaps and reconcile divergent lines of reasoning. 
By requiring agents to critique others' rationales, cross-review improves accuracy and exposes intermediate reasoning in a form that OCEs can audit.
After each agent produces an initial answer with stepwise reasoning, the \textit{Orchestrator} initiates cross-review by bundling each answer and its rationale and dispatching it to the other two agents for peer review (\eg $\mathcal{A}_1$ reviews $\mathcal{A}_2$ and $\mathcal{A}_3$, and symmetrically for the others). 
Each agent then returns concise \textit{review advice} to its peers, focusing on overlooked or weak evidence, unclear reasoning that warrants clarification, and plausible alternative hypotheses. 
Then, the \textit{Orchestrator} prompts agents to refine their answers in accordance with the \textit{review advice}.
All review messages and refined rationales are recorded and included in the \textit{Root Cause Report}, creating an auditable trail that substantiates the final diagnosis.

\subsubsection{Online Deployment and Offline Training}
The \textit{Root Cause Report} serves two purposes: guiding operational remediation and providing learning feedback for continual improvement.
In online deployment, engineers validate the report via mitigation actions. 
A successful mitigation confirms the diagnosis and closes the incident, whereas a failed mitigation triggers system re-analysis.
The \textit{Orchestrator} augments the original \textit{Query} with feedback indicating that the previous diagnosis was incorrect, then re-invokes the multi-agent collaboration until a diagnosis is confirmed by successful mitigation. 
In offline training, accumulated reports are fed to the self-evolution mechanism (Section~\ref{sec:self-evolution}) to update agents via PPO-guided optimization and to distill reusable experience. 
This closes the loop between operations and learning, improving diagnostic capability over time while preserving deployability.

\subsection{Self-evolution Mechanism}\label{sec:self-evolution}

\begin{figure}[htbp]
    \centering
    \includegraphics[width=\linewidth]{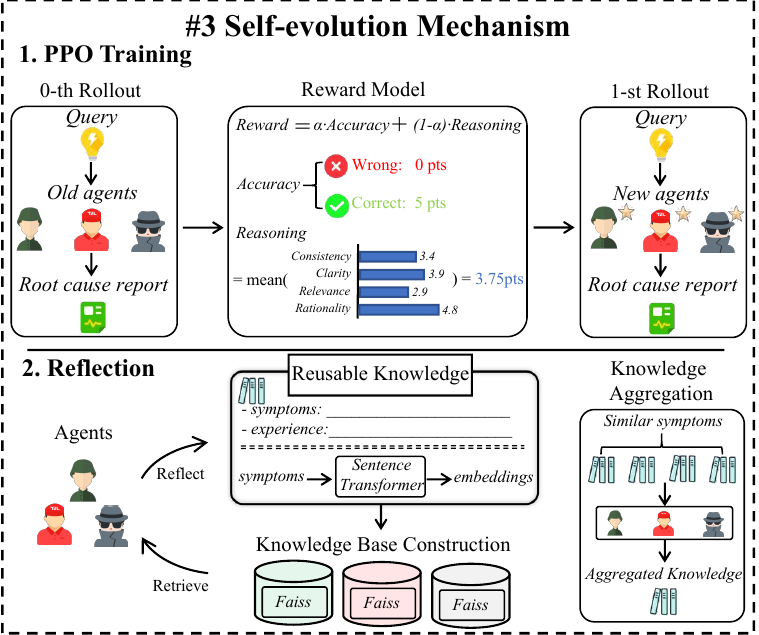}
    \caption{
    \textbf{Self-evolution Mechanism}. Internally, agents are fine-tuned via PPO training with a carefully designed reward model (top). Externally, a reflection process distills reusable knowledge into a task-specific knowledge base, which is later leveraged through RAG for knowledge injection (bottom).
    }
    \label{fig:self-evolution}
\end{figure}

To continuously enhance the causal diagnostic capability of \name, we design a self-evolution mechanism that operates from two complementary perspectives as illustrated in Fig.~\ref{fig:self-evolution}. 
Internally, agents are updated through PPO-based fine-tuning with well-designed rewards, thereby strengthening their intrinsic reasoning ability.
Externally, \name invokes a reflection process in which agents summarize past diagnostic experiences and distill reusable knowledge into a knowledge base. 
In this manner, the system is able to evolve beyond static pattern memorization, progressively improving its diagnostic performance from real-world cases, akin to how OCEs accumulate expertise over time.

\subsubsection{PPO Training}
To enhance the diagnostic reasoning capability of \name, we adopt PPO \cite{ppo}, a stable reinforcement learning algorithm that aligns well with capability-centric learning paradigms, to train the three expert agents, $\mathcal{A}_1$, $\mathcal{A}_2$, and $\mathcal{A}_3$.
PPO's clipped objective and an adaptive KL penalty help stabilize learning and limit excessive policy shifts while maintaining sample efficiency.

A carefully designed reward model is employed to guide the optimization process. 
Optimizing only for accuracy can encourage shortcut behavior, whereas incorporating stepwise reasoning quality into the reward improves auditability and strengthens peer critique in cross-review.
We therefore design the reward model to combine accuracy and reasoning quality as complementary signals for PPO training.
Accuracy is measured in a binary manner, assigning a score of 5 for a correct diagnosis and 0 otherwise. 
Reasoning quality is assessed by an external judge model \textit{Qwen3-235B-A22B}, which is well suited for this role since evaluating reasoning evidence is more tractable than generating it \cite{g-eval}.
As it is used only for reward estimation rather than as the reasoning backbone, this design decouples reward evaluation from the policy model while incurring only limited additional cost.
It scores each reasoning chain along four dimensions—consistency, clarity, relevance, and rationality—on a 0–5 scale. 
The average of these scores is then combined with the accuracy reward, weighted by a tunable coefficient $\alpha \in [0,1]$, allowing flexible adjustment of their relative importance.
During training, each rollout, which refers to the reasoning trajectory produced by \name for a given \textit{Query}, is assigned a reward according to the above scheme. 
The reward is then used to update the parameters of the three expert agents individually via PPO, ensuring that each agent improves its task-specific reasoning competence.

\subsubsection{Reflection}
While internal parameter optimization via PPO training enhances task-specific reasoning capability, it is insufficient for sustaining long-term evolution. 
In practice, diagnostic systems must not only adapt their parameters but also explicitly accumulate reusable experience, much like OCEs who distill repeated operational experience into shared troubleshooting guides.
To this end, \name incorporates a reflection mechanism that distills knowledge from successfully resolved cases, ensuring that only validated trajectories contribute to future reasoning. 
After completing a diagnostic task correctly, the corresponding agent reflects on its reasoning trajectory and outcome, abstracting reusable knowledge, such as characteristic symptom–root cause patterns. 
Each knowledge entry is structured as a pair $<symptoms, experience>$, where the symptoms serve as the key. 
The entries are embedded using a Sentence-Transformer \cite{sentence-bert} and stored in a faiss vector database.
Importantly, each expert agent (\ie $\mathcal{A}_1$, $\mathcal{A}_2$, $\mathcal{A}_3$) maintains its own knowledge base, as the nature of accumulated experience differs across diagnostic tasks.
By grounding reflection in successfully verified cases, the system avoids propagating erroneous reasoning and builds a reliable knowledge repository that complements PPO training.

When handling a new case, the agent first generates a symptom key based on the input data descriptions and then queries its task-specific knowledge base via RAG, injecting the retrieved experiences into its prompt as auxiliary context.
To keep the knowledge base compact and reliable, we aggregate entries that share the same symptom key: if two entries conflict, the newer replaces the older to account for obsolete practices; if they are complementary, we merge them so that multiple valid experiences can coexist. 
This reconciliation preserves conciseness without discarding useful diversity. 
During early deployment, the knowledge base may be sparse, and if retrieval fails, the agent falls back to pure CoT reasoning without external retrieval to maintain robustness \cite{lewis2020retrieval}.

Together, PPO-based training and reflection provide complementary paths for continual improvement: the former strengthens agents' intrinsic reasoning strategies, while the latter consolidates reusable knowledge from past cases. 
By combining implicit parameter updates with explicit experience accumulation, \name evolves beyond static pattern memorization and continually enhances its diagnostic expertise. 
This evolution addresses (\textbf{\textit{C3}}), enabling sustained capability growth in MAS-based IM and mirroring how OCEs refine their skills through both practice and knowledge accumulation.

\section{Experiments}

\begin{table*}[htbp]
\centering
\caption{Comparison with baselines across seed LLMs. Metrics (\%): Correct and Partial are reported per system and as averages; the final column reports average time per incident (s/case).}
\small
\setlength{\tabcolsep}{6pt}
\renewcommand{\arraystretch}{1.25}
\begin{tabularx}{\textwidth}{l l | *{9}{Y}}
\toprule
\multirow{2}{*}{Seed LLM} & \multirow{2}{*}{Method} & 
\multicolumn{2}{c|}{Telecom} & \multicolumn{2}{c|}{Bank} & 
\multicolumn{2}{c|}{Market} & \multicolumn{2}{c}{Avg} & 
\multirow{2}{*}{s/case} \\
\cmidrule(lr){3-10}
 & & Correct & Partial & Correct & Partial & Correct & Partial & Correct & Partial & \\
\midrule
\multirow{4}{*}{Qwen2.5-14B-Instruct-1M}    
 & \name                         & 30.00 & 45.00 & \textbf{18.52} & \textbf{40.74} & \textbf{10.17} & \textbf{40.68} & \textbf{16.54} & \textbf{41.35} & 71.18 \\
 & CoT \cite{cot}                & 10.00 & 20.00 & 0.00 & 1.85 & 5.08 & 13.55 & 1.50 & 9.77 & 16.83 \\
 & ReAct \cite{react}            & 5.00  & 20.00 & 0.00 & 1.85 & 0.00 & 5.08  & 0.75 & 6.01 & 164.24 \\
 & Reflexion \cite{reflexion}    & 0.00  & 10.00 & 3.70 & 11.11 & 1.69 & 5.08  & 2.26 & 8.27 & 241.55 \\
\midrule
\multirow{4}{*}{gpt-oss-20b}   
 & \name                         & \textbf{40.00} & \textbf{55.00} & 5.56 & 11.11 & 6.78 & 16.95 & 11.28 & 20.30 & 53.48 \\
 & CoT \cite{cot}                & 0.00  & 5.00  & 0.00 & 0.00  & 0.00 & 0.00  & 0.00 & 0.75 & 27.42 \\
 & ReAct \cite{react}            & 0.00  & 0.00  & 0.00 & 0.00  & 1.69 & 1.69  & 0.75 & 0.75 & 81.09 \\
 & Reflexion \cite{reflexion}    & 0.00  & 0.00  & 0.00 & 0.00  & 1.69 & 3.38  & 0.75 & 1.50 & 165.74 \\
\midrule
\multirow{4}{*}{Phi-3-medium-128k-instruct} 
 & \name                         & 10.00 & 35.00 & 0.00 & 9.26  & 3.39 & 10.17 & 3.01 & 13.53 & 57.94 \\
 & CoT \cite{cot}                & 0.00  & 10.00 & 0.00 & 0.00  & 0.00 & 0.00  & 0.00 & 1.50 & 20.33 \\
 & ReAct \cite{react}            & 0.00  & 0.00  & 0.00 & 0.00  & 0.00 & 0.00  & 0.00 & 0.00 & 48.42 \\
 & Reflexion \cite{reflexion}    & 0.00  & 0.00  & 0.00 & 0.00  & 0.00 & 0.00  & 0.00 & 0.00 & 100.05 \\
\midrule
Claude 3.5 Sonnet & RCA-Agent \cite{openrca} & 20.00 & 35.00 & 16.67 & 35.19 & 3.39 & 28.81 & 11.28 & 32.33 & 287.71 \\
\midrule
\multicolumn{1}{l}{} & ART \cite{art}           & 0.00  & 15.00 & 0.00  & 11.11 & 1.69 & 23.73 & 0.75  & 17.29 & \textbf{2.27} \\
\bottomrule
\end{tabularx}
\label{tab:baseline}
\end{table*}

\subsection{Experimental Setup}
\subsubsection{Datasets}
We conducted extensive experiments on the OPENRCA \cite{openrca} benchmark dataset which consists of 335 incident cases collected from 3 heterogeneous microservice systems (\ie Telecom, Bank, Market) deployed in the real-world environment, accompanied by over 68 GB of telemetry data.
The dataset underwent rigorous preprocessing, including standardization and balancing, and was further calibrated by 3 experienced engineers who verified that each root cause label could be supported by the associated telemetry, ensuring the reliability of the benchmark. 
Specifically, (1) \textbf{Telecom} includes 5 failure types in $\mathbb{C}$ and 43 candidate components in $\mathbb{R}$, comprising 22 virtual machine operating systems, 8 pods, and 13 database services; (2) \textbf{Bank} contains 8 failure types and 14 candidate components corresponding to its 14 pods; (3) \textbf{Market} has 15 failure types and 56 candidate components, consisting of 6 nodes, 40 pods, and 10 services. 

\subsubsection{Baselines}
We compare \name against five frameworks: two designed specifically for IM (\ie ART \cite{art} and RCA-Agent \cite{openrca}), and three general-purpose open-source frameworks (\ie CoT \cite{cot}, ReAct \cite{react_yao}, Reflexion \cite{reflexion}).
For DL-based IM approaches, we select ART \cite{art} due to its state-of-the-art performance, while other methods such as DeepHunt \cite{deephunt} and DiagFusion \cite{diagfusion} are excluded as they do not cover all tasks required in our scenario (\eg AD).
For LLM-based IM approaches, we include RCA-Agent \cite{openrca}, the official OPENRCA diagnostic framework and the top-performing approach on its benchmark. 
We evaluate it with \textit{Claude 3.5 Sonnet} as the seed LLM to follow its strongest configuration, since RCA-Agent becomes ineffective when instantiated with $\le$20B-scale models due to its strong dependence on code generation and error handling capabilities.
We exclude other LLM-based IM methods, specifically mABC \cite{mabc}, ICL\_RCA \cite{icl_rca}, COMET \cite{comet}, due to their lack of coverage for all required tasks in our scenario, and RCACopilot \cite{rcacopilot}, as it is not an open-source framework. 
To ensure a comprehensive and fair comparison, we also incorporate three well-known general-purpose open-source frameworks—CoT \cite{cot}, ReAct \cite{react_yao}, and Reflexion \cite{reflexion}—all of which leverage LLMs for reasoning and decision-making.
For the seed LLMs, we restrict to models with at most 20B parameters and select \textit{Qwen2.5-14B-Instruct-1M}, \textit{gpt-oss-20b}, and \textit{Phi-3-medium-128k-instruct}.

\subsubsection{Evaluation Metrics}
As described in Section \ref{sec:problem_definition}, our system needs to handle arbitrary combinations of three subtasks: AD, FT, and RCL.
Each subtask follows a clearly defined success criterion. 
An AD subtask is considered correct if the predicted timestamp lies within one minute ($\pm60$ seconds) of the ground-truth label. 
FT and RCL are evaluated in a single-pass setting: a prediction is counted as correct only when it exactly matches the ground-truth label (Top-1 match).
Then, for any input query that may combine multiple subtasks, we utilize 
$Correct = \frac{num_c}{N}$ and $Partial = \frac{num_p}{N}$
as evaluation metrics, where \(N\) denotes the total number of queries in the dataset, \(num_c\) is the number of queries for which every requested subtask is correctly solved, and \(num_p\) is the number of queries for which at least one requested subtask is correctly solved.

\subsubsection{Implementations}
We implemented \name using Python 3.10.16 with PyTorch 2.6.0, Transformers 4.51.1, and accelerate 1.7.0. 
We perform a random split, assigning 60\% of the data to the training set and the remainder to the test set.
Experiments were conducted on a server with 16-core Intel Xeon Gold 5416S CPU, 376GB RAM, and 8 GPUs (48GB memory each).
To ensure result reliability, we repeated each experiment five times and reported the average performance.

\subsection{Performance Evaluation}

Table~\ref{tab:baseline} presents the performance of \name and baselines on the OPENRCA \cite{openrca} benchmark across three microservice systems.
\name consistently attains the best average scores on both Correct and Partial, surpassing the SOTA by \textbf{46.63\%} in Correct and \textbf{27.90\%} in Partial.
General-purpose open-source prompting frameworks (\ie CoT \cite{cot}, ReAct \cite{react}, Reflexion \cite{reflexion}) perform unevenly, as IM is demanding even for experienced OCEs, and approaches not tailored to the task struggle with heterogeneous observability data and tightly coupled causal relations. 
CoT \cite{cot} uses step-wise reasoning and tends to be stable, which is why we also introduce CoT prompting in our design.
ReAct \cite{react} integrates tool use, but coordinating tools over diverse observations poses great challenges in planning.
Reflexion \cite{reflexion} builds on ReAct by reflecting on prior reasoning paths and yields some improvements, but the gains are limited because ReAct often produces noisy and disorganized trajectories.
Among LLM-based baselines, RCA-Agent \cite{openrca} ranks second on average Correct/Partial by introducing an executor agent that synthesizes and runs programs. 
However, this design relies heavily on the error-handling and planning capacity of a closed-source LLM (Claude 3.5 Sonnet).
This dependency incurs substantial inference costs and poses privacy-exposure risks when sensitive observability and incident data are sent to third-party APIs.
The DL-based baseline ART \cite{art} achieves relatively acceptable performance as it fits the system-specific data patterns with its well-designed architecture.
At the same time, we observe significant sensitivity to the underlying backbone: as the seed LLM changes, performance varies for both \name and the general-purpose baselines, yet under every seed LLM \name achieves the best Correct/Partial averages.

To quantify runtime cost, we report the average time per incident case in the final column of Table~\ref{tab:baseline}. 
\name processes a case in roughly 1 minute on average, demonstrating its high efficiency while maintaining strong performance.
Among general-purpose open-source prompting frameworks, CoT \cite{cot} completes a case in about 20 seconds but with much lower accuracy, ReAct \cite{react} slows due to multi-round tool calls, and Reflexion \cite{reflexion} is even slower as it reflects upon prior trajectories of ReAcT. 
RCA-Agent \cite{openrca} takes close to 5 minutes per case because its executor frequently regenerates and reruns code when errors occur. 
ART \cite{art} is the fastest due to lightweight neural networks and attains moderate accuracy. 
Overall, \name achieves a balanced profile across efficiency and accuracy, while operating on a local 14B model that avoids the high costs associated with closed-source APIs.

\subsection{Ablation Study}
To validate the contribution of \name's core modules, we conduct an ablation study under different conditions: \textbf{A1}: without data processor, \textbf{A2}: without cross-review mechanism, \textbf{A3}: without reflection, \textbf{A4}: without PPO fine-tuning. 
As shown in Table~\ref{tab:ablation}, \name (trained with 60\% cases) outperforms all the variants and the results reveal three key findings:
(1) \textbf{Textual data descriptions are necessary for LLM reasoning (A1)}. 
For A1, we still filter the anomalous observability data to fit the context window, but feed them in their original form rather than as structured textual descriptions. 
We observed that LLMs struggle to deal with massive numerical inputs, leading to unstable reasoning, while structured textual descriptions expose salient cues and enable more reliable inference.
(2) \textbf{Cross-review is critical for robust diagnosis under complex IM scenarios (A2)}.
For A2, we remove the cross-review mechanism and let the three expert agents generate their diagnoses independently without reviewing or refining each other's outputs.
Both Correct and Partial drop sharply, indicating that cross-review is essential because it incorporates complementary perspectives, reconciles divergent reasoning, and uncovers missing or weak evidence across agents, all of which are crucial for maintaining diagnostic quality.
(3) \textbf{Internal training and external reflection are complementary for sustained gains (A3, A4)}.
For A3, we fine-tune the agents with PPO only, whereas for A4 we disable PPO and rely solely on reflection-based knowledge distillation with RAG at inference.
Both variants underperform the full model, with lower Correct/Partial across systems, indicating that both internal training and external reflection to be indispensable for improving performance, and in combination they deliver complementary gains as the system evolves.

\begin{table}[htbp]
\centering
\caption{Ablation study on key components (\textbf{Seed LLM: Qwen2.5-14B-Instruct-1M}). Results are dataset-averaged Correct/Partial (\%).}
\small
\setlength{\tabcolsep}{8pt}
\renewcommand{\arraystretch}{1.25}
\begin{tabularx}{0.48\textwidth}{l|YY}
\toprule
\multirow{2}{*}{Variant} & \multicolumn{2}{c}{Average} \\
\cmidrule(lr){2-3}
 & Correct & Partial \\
\midrule
A1: w/o Data Processor  & 2.26 & 6.02 \\
A2: w/o Cross-review   & 6.77 & 21.80 \\
A3: w/o Reflection      & 10.53 & 26.32 \\
A4: w/o PPO             & 12.78 & 33.08 \\
\name\ (0\%)             & 8.27 & 27.07 \\
\name\ (60\%)              & \textbf{16.54} & \textbf{41.35} \\
\bottomrule
\end{tabularx}
\vskip3pt 
\parbox{0.48\textwidth}{
\footnotesize
\textbf{Notes:} Percentages indicate the fraction of incident cases used for the self-evolution mechanism (PPO and reflection). ``\name\ (0\%)'' disables self-evolution, ``\name\ (60\%)'' applies self-evolution using 60\% of the cases.
}
\label{tab:ablation}
\end{table}

\begin{table}[h]
\centering
\caption{Self-evolution capability under different training budgets.}
\small
\setlength{\tabcolsep}{3pt}
\renewcommand{\arraystretch}{1.15}
\resizebox{\columnwidth}{!}{
\begin{tabularx}{\columnwidth}{l c | *{2}{Y}}
\toprule
\multirow{2}{*}{Seed LLM} & \multirow{2}{*}{Training (\%)} & \multicolumn{2}{c}{Avg}\\
\cmidrule(lr){3-4}
 &  & Correct & Partial\\
\midrule
\multirow{4}{*}{Qwen2.5-14B-Instruct-1M}
 & 0  & 8.27 & 27.06 \\
 & 20 & 12.03 & 33.83 \\
 & 40 & 14.29 & 38.35 \\
 & 60 & \textbf{16.54} & \textbf{41.35} \\
\bottomrule
\end{tabularx}
}
\label{tab:self_evo}
\end{table}

\subsection{Self-evolution Evaluation}
We further evaluate self-evolution to verify continual capability growth. 
We vary the number of training cases used for updates after deployment and always report Correct and Partial on a held-out test set. 
As shown in Table~\ref{tab:self_evo}, \name exhibits a steady increase as more incidents are processed. 
The gains arise from two complementary sources: internal parameter updates via PPO that align the agents with the reward signal, and external reflection that distills experience into a knowledge base for retrieval during inference. 
This pattern mirrors how OCEs accumulate expertise through practice.
The results validate the self-evolution mechanism and support \name's suitability for long-term deployment.

\section{Discussion}
\subsection{Deployment}
We deployed \name as a real-time incident diagnostic system in Lenovo's production environment, using \textit{Qwen2.5-14B-Instruct-1M} as the seed LLM.
The environment spans on the order of $2.5\times10^4$ infrastructure instances across data-center facilities, hardware, virtualization/cloud platforms, middleware, databases, and application/service layers.
Its observability stack includes \textit{VictoriaMetrics} for metrics and \textit{OpenTelemetry} for logs and traces.
Over 53 days, \name processed 10,492 incidents in production environment.

\textbf{Effectiveness.}
Over the 53-day deployment, \name achieved an overall diagnostic accuracy of 84.09\%.
Here, accuracy is defined by whether the final validated report correctly identifies the root-cause reason and component, as verified through OCE-confirmed mitigation.
This result is higher than those in Table~\ref{tab:baseline} because production queries provide richer incident context (\eg alert information) than OPENRCA \cite{openrca}, which narrows the search space and improves diagnostic fidelity.
In traditional operations, resolving one incident typically required the collaboration of 3 experienced OCEs for about 2.5 hours, whereas \name reduced to 126 seconds.
Routine and localized incidents accounted for roughly 70\% of all incidents, mainly involving recurring types like \textit{Disk Space Exhaustion}, and \textit{Proxy Misconfiguration}.
For these incidents, \name achieved 97\% accuracy with an average diagnosis time of only 30 seconds, demonstrating high efficiency and effectiveness.
The remaining incidents characterized by cross-component interactions, ambiguous symptoms, or incomplete evidence were challenging, for which \name was allowed to perform up to three rounds of re-analysis yet still achieved only 54\% accuracy.
Nevertheless, even when the diagnosis was wrong, the generated \textit{Root Cause Report} still provided OCEs with useful evidence and plausible hypotheses, eliminating the need for purely manual investigation over massive observability data.
Therefore, \name still substantially reduced the effort required from OCEs and accelerated incident mitigation in difficult cases.
Importantly, we observed clear benefits from self-evolution during deployment.
Some complex incidents that initially required multiple re-analysis rounds could later be diagnosed correctly in a single attempt, and some failure types that were initially unresolved became diagnosable when they reappeared.
This trend indicates that \name continuously strengthens its diagnostic capability after deployment, especially for recurring complex incident patterns.

\subsection{Case Study}

\begin{figure}[htbp]
    \centering
    \includegraphics[width=\linewidth]{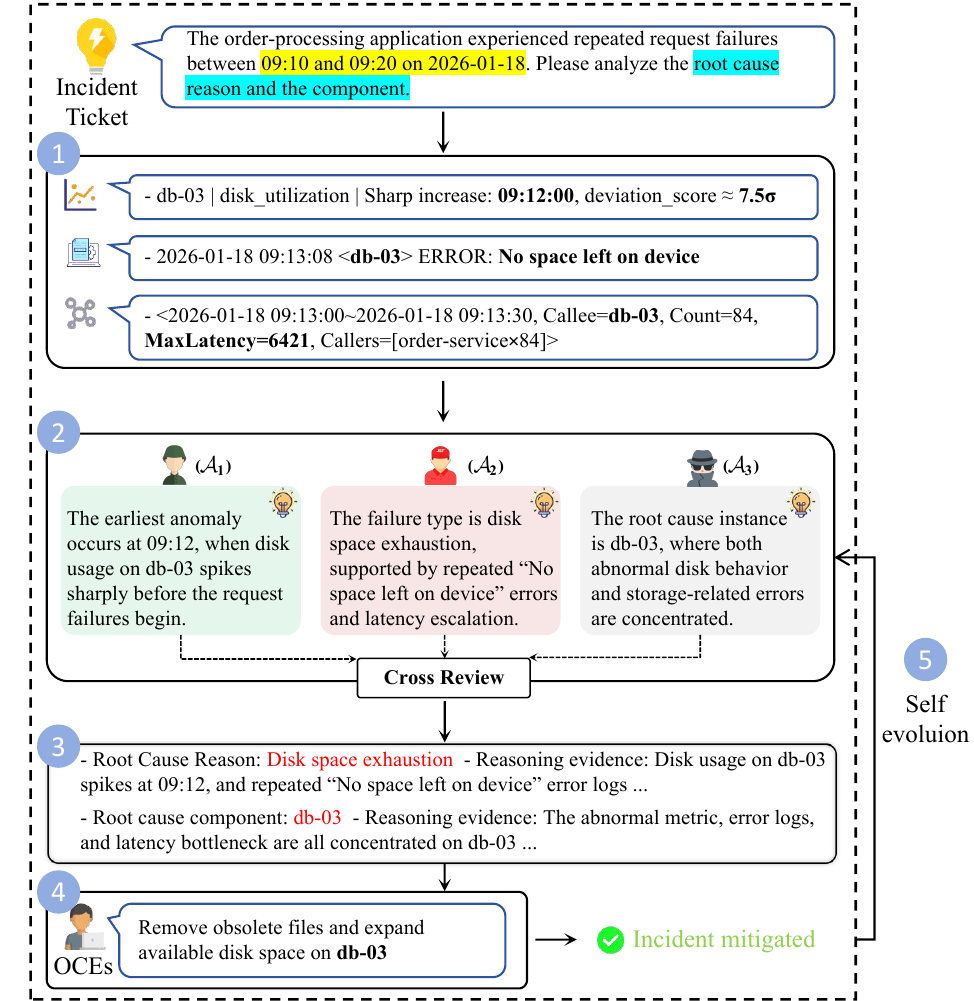}
    \caption{An illustrative case.
    }
    \label{fig:case-study}
    \vspace{-3mm}
\end{figure}

To illustrate how \name operates after deployment, we present a representative case study from Lenovo's production environment. 
All service names and incident details have been anonymized for security and privacy reasons.
As shown in Fig.~\ref{fig:case-study}, the diagnostic process began when an incident ticket was automatically generated for an order-processing application, reporting repeated request failures between 09:10 and 09:20. 
(1) After interpreting the ticket, \name retrieved multimodal observations within this time range and converted them into unified textual descriptions for analysis. 
In this case, it identified a sharp increase in disk utilization on instance \texttt{db-03}, repeated ``No space left on device'' errors in the logs of \texttt{db-03}, and high-latency spans concentrated on calls from \texttt{order-service} to \texttt{db-03}. 
(2) Based on these evidences, the anomaly detection agent inferred that the earliest abnormal timestamp was 09:12, the failure triage agent concluded that the incident was caused by disk space exhaustion, and the root cause localization agent identified \texttt{db-03} as the faulty instance. 
(3) During cross-review, the failure triage agent agreed with the localization result because the storage-related log errors and trace bottlenecks both pointed to \texttt{db-03}; the anomaly detection agent further confirmed that the disk-utilization surge on \texttt{db-03} appeared before the request failures; and the localization agent ruled out \texttt{order-service} itself because its latency increase occurred only after calls to \texttt{db-03} became slow. 
After integrating these consistent reviews, \name generated a final \textit{Root Cause Report}, identifying disk space exhaustion on \texttt{db-03} as the root cause. 
(4) The report was delivered to the frontline OCE, who cleaned obsolete files and expanded the available disk space on \texttt{db-03}, after which the service recovered successfully. 
(5) Since the mitigation validated the diagnosis, the confirmed case was fed back to \name for self-evolution, where it was used to improve the agents and distill reusable experience for future similar incidents.

\subsection{Interpretability in Deployment}

\begin{figure}[htbp]
    \centering
    \includegraphics[width=\linewidth]{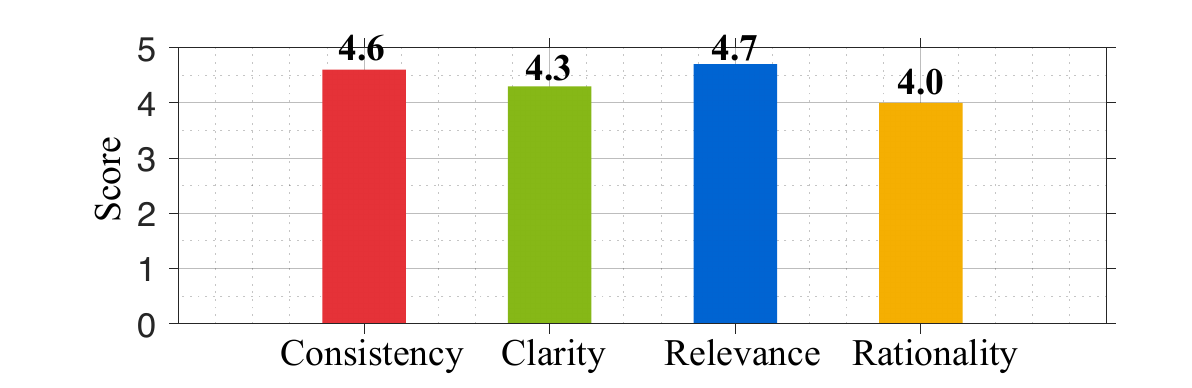}
    \caption{\textbf{Interpretability assessment in deployment.}
    Mean scores given by frontline OCEs on four dimensions.}
    \label{fig:interp}
    \vspace{-3mm}
\end{figure}

To understand whether \name's reports are practically useful after deployment, we asked three frontline OCEs who use the system in production to assess its interpretability. 
Specifically, we randomly sampled 150 incidents from the 10,492 incidents, and the three OCEs rated the generated root cause reports on four dimensions, \ie \textit{consistency}, \textit{clarity}, \textit{relevance}, and \textit{rationality}, using a 1-5 scale.
As shown in Fig.~\ref{fig:interp}, \name receives consistently favorable ratings overall, with mean scores of 4.6, 4.3, 4.7, and 4.0, respectively. 
These results suggest that the reports are generally coherent, easy to follow, and well aligned with the incident context. 
Overall, they demonstrate that \name provides strong interpretability in real deployment and serves as a practical diagnostic assistant for frontline OCEs.

\subsection{Lessons Learned}

\subsubsection{Diagnostic Interpretability}
After deployment, we found that a diagnosis alone was often insufficient for OCEs to trust and act upon the system's output. 
In practice, OCEs need clear evidence and a traceable reasoning process to understand why a conclusion is reached.
We therefore observed that interpretability is not merely a desirable property, but a prerequisite for practical adoption: when the diagnosis is correct, it increases trust and actionability; even when it is incorrect, the exposed reasoning process and collected evidence can still substantially assist OCEs in incident mitigation.

\subsubsection{Self-Feedback for Sustainability}
We also learned that static IM approaches are difficult to sustain in dynamic production environments, where previously unseen incident types continuously emerge.
Without an effective feedback mechanism, diagnostic performance can gradually degrade as the environment evolves.
By feeding validated deployment outcomes back into \name for self-evolution, the system gradually adapts to new failure patterns and reduces the need for repeated manual maintenance, which is critical for long-term deployability.

\section{Related~work}
\subsection{DL-based IM approaches.}
DL-based approaches leverage neural networks to extract failure patterns from observability data and predict root causes, employing either supervised \cite{diagfusion,eadro} or self-supervised \cite{art,deephunt} learning.
Early studies focused on unimodal data sources, such as metrics \cite{automap,coral,patternmatcher,dejavu,cloudranger,ms-rank}, logs \cite{swisslog,logkg,logm,onion}, or traces \cite{mepfl,tracerank,tracerca,tracenet}, which limited their ability to capture a comprehensive view of system states. 
Recent advances have adopted multimodal fusion, exemplified by ART \cite{art}, DeepHunt \cite{deephunt}, and DiagFusion \cite{diagfusion}, which integrate heterogeneous observability data to improve diagnostic accuracy and efficiency. 
Despite these improvements, DL-based IM methods face three key limitations. 
First, models often overfit system-specific patterns, leading to poor generalization across heterogeneous and evolving microservice systems. 
Second, their black-box nature hinders interpretability, offering little insight into the reasoning process behind root cause predictions, which reduces trustworthiness for OCEs. 
Finally, these models are trained in a static manner and cannot accumulate diagnostic experience over time, making them ill-suited for sustained deployment in dynamic microservice environments. 

\begin{table}[t]
\centering
\caption{Comparison of IM categories. }
\small
\setlength{\tabcolsep}{8pt}
\renewcommand{\arraystretch}{1.25}

\begin{tabularx}{0.48\textwidth}{l|YYYY}
\toprule
\textbf{Categories} &
\textbf{Gen?} &
\textbf{Int?} &
\textbf{Cos?} &
\textbf{Evo?} \\
\midrule
DL-based \cite{art,deephunt,diagfusion}            & \pbEmpty & \pbEmpty & \pbFull & \pbEmpty \\
LLM-based \cite{rcacopilot,icl_rca,ahmed}       & \pbHalf & \pbHalf & \pbEmpty  & \pbEmpty  \\
MAS-based \cite{mabc,trioxpert,dbot}             & \pbHalf & \pbFull  & \pbFull & \pbEmpty  \\
\name        & \pbFull & \pbFull  & \pbFull  & \pbFull  \\
\bottomrule
\end{tabularx}
\vskip 3pt 
\parbox{0.48\textwidth}{
\footnotesize
\textbf{Notes:}
``Gen?'': Generalizable? ``Int?'': Interpretable? ``Cos?'': Cost-Efficient? ``Evo?'': Self-Evolving? ``\pbFull'': full support, ``\pbHalf'': partial support, ``\pbEmpty'': no support. 
}
\label{tab:agent-compare}
\vspace{-5.3mm}
\end{table}

\subsection{LLM-based IM approaches.} 
With the rapid progress of LLMs, recent studies have begun to explore their potential in IM. 
Ahmed \etal \cite{ahmed} fine-tuned GPT-3.X on domain corpora for direct root cause prediction from incident titles and summaries. 
RCACopilot \cite{rcacopilot} and Zhang \etal \cite{icl_rca} use in-context learning, retrieving semantically similar past incidents as examples for few-shot prediction.
While these approaches demonstrate the feasibility of LLMs for IM, they also exhibit critical limitations. 
First, most methods rely on surface-level similarity matching or prompt engineering, rather than performing grounded reasoning over raw observability data, which weakens alignment with actual system behavior. 
Second, their outputs typically lack transparent reasoning chains, offering little interpretability or auditability for OCEs in high-stakes operational scenarios. 
Finally, they are often built on closed-source proprietary models such as GPT-4, resulting in high inference costs and strong dependency on external APIs, which poses substantial barriers to cost-efficient and trustworthy deployment in real-world environments. 
Building on LLMs, recent efforts have explored multi-agent system (MAS)-based IM, leveraging specialized agents for collaborative reasoning. 
For instance, mABC \cite{mabc} adopts a blockchain-inspired decentralized voting mechanism among LLM-based agents for collective root cause inference; TrioXpert \cite{trioxpert} proposes a multi-expert architecture integrating modality-specific preprocessing and agent collaboration for IM; D-Bot \cite{dbot} focuses on database diagnosis, leveraging agent collaboration with specialized roles (\eg CPU, Network) to generate timely diagnostic reports.
In parallel, OPENRCA \cite{openrca} introduces the first benchmark for natural-language-driven IM.
Notably, even top methods using closed-source LLMs (\eg Claude 3.5 Sonnet, GPT-4o, Gemini 1.5 Pro) only achieve 11.34\% accuracy, highlighting IM's inherent difficulty and current approach limitations.

Overall, existing IM methods lack generalization, interpretability, cost-efficiency and self-evolution (Table~\ref {tab:agent-compare}), while \name uniquely unifies these capabilities, enabling long-term deployment in real-world microservice environments.

\section{Conclusion}
This work introduces \name, a lightweight, self-evolving multi-agent system for IM. 
We incorporate a training-free data processor to handle massive, heterogeneous observability data, a multi-agent collaboration framework that renders diagnostic inference transparent and auditable, and a dual self-evolution mechanism integrating internal model updates with external experience accumulation. 
With this design, \name is generalizable, interpretable, cost-efficient, and self-evolving, making it a practically deployable and sustainable solution for real-world microservice systems.
We believe that the concept of constructing a multi-agent system with a well-designed data processor, a multi-agent collaboration framework, and a self-evolution mechanism can generalize to other complex scenarios that involve massive, heterogeneous data.

\section{Data Availability Statement}
All code, prompts and data are available in the repo: \url{https://anonymous.4open.science/r/OpsAgent-CCC0}.


\newpage

\bibliographystyle{ACM-Reference-Format}
\bibliography{main.bib}


\begin{thebibliography}{58}


\ifx \showCODEN    \undefined \def \showCODEN     #1{\unskip}     \fi
\ifx \showISBNx    \undefined \def \showISBNx     #1{\unskip}     \fi
\ifx \showISBNxiii \undefined \def \showISBNxiii  #1{\unskip}     \fi
\ifx \showISSN     \undefined \def \showISSN      #1{\unskip}     \fi
\ifx \showLCCN     \undefined \def \showLCCN      #1{\unskip}     \fi
\ifx \shownote     \undefined \def \shownote      #1{#1}          \fi
\ifx \showarticletitle \undefined \def \showarticletitle #1{#1}   \fi
\ifx \showURL      \undefined \def \showURL       {\relax}        \fi
\providecommand\bibfield[2]{#2}
\providecommand\bibinfo[2]{#2}
\providecommand\natexlab[1]{#1}
\providecommand\showeprint[2][]{arXiv:#2}

\bibitem[goo(2025)]%
        {google_crash}
 \bibinfo{year}{2025}\natexlab{}.
\newblock \bibinfo{title}{Incident Report of google cloud outage}.
\newblock
  \bibinfo{howpublished}{\url{https://status.cloud.google.com/incidents/ow5i3PPK96RduMcb1SsW}}.
\newblock


\bibitem[Achiam et~al\mbox{.}(2023)]%
        {gpt4}
\bibfield{author}{\bibinfo{person}{Josh Achiam}, \bibinfo{person}{Steven
  Adler}, \bibinfo{person}{Sandhini Agarwal}, \bibinfo{person}{Lama Ahmad},
  \bibinfo{person}{Ilge Akkaya}, \bibinfo{person}{Florencia~Leoni Aleman},
  \bibinfo{person}{Diogo Almeida}, \bibinfo{person}{Janko Altenschmidt},
  \bibinfo{person}{Sam Altman}, \bibinfo{person}{Shyamal Anadkat},
  {et~al\mbox{.}}} \bibinfo{year}{2023}\natexlab{}.
\newblock \showarticletitle{Gpt-4 technical report}.
\newblock \bibinfo{journal}{\emph{arXiv preprint arXiv:2303.08774}}
  (\bibinfo{year}{2023}).
\newblock


\bibitem[Ahmed et~al\mbox{.}(2023)]%
        {ahmed}
\bibfield{author}{\bibinfo{person}{Toufique Ahmed}, \bibinfo{person}{Supriyo
  Ghosh}, \bibinfo{person}{Chetan Bansal}, \bibinfo{person}{Thomas Zimmermann},
  \bibinfo{person}{Xuchao Zhang}, {and} \bibinfo{person}{Saravan Rajmohan}.}
  \bibinfo{year}{2023}\natexlab{}.
\newblock \showarticletitle{Recommending root-cause and mitigation steps for
  cloud incidents using large language models}. In
  \bibinfo{booktitle}{\emph{2023 IEEE/ACM 45th International Conference on
  Software Engineering (ICSE)}}. IEEE, \bibinfo{pages}{1737--1749}.
\newblock


\bibitem[Ataei et~al\mbox{.}(2025)]%
        {elicitron}
\bibfield{author}{\bibinfo{person}{Mohammadmehdi Ataei},
  \bibinfo{person}{Hyunmin Cheong}, \bibinfo{person}{Daniele Grandi},
  \bibinfo{person}{Ye Wang}, \bibinfo{person}{Nigel Morris}, {and}
  \bibinfo{person}{Alexander Tessier}.} \bibinfo{year}{2025}\natexlab{}.
\newblock \showarticletitle{Elicitron: A large language model agent-based
  simulation framework for design requirements elicitation}.
\newblock \bibinfo{journal}{\emph{Journal of Computing and Information Science
  in Engineering}} \bibinfo{volume}{25}, \bibinfo{number}{2}
  (\bibinfo{year}{2025}), \bibinfo{pages}{021012}.
\newblock


\bibitem[Chen et~al\mbox{.}(2024)]%
        {rcacopilot}
\bibfield{author}{\bibinfo{person}{Yinfang Chen}, \bibinfo{person}{Huaibing
  Xie}, \bibinfo{person}{Minghua Ma}, \bibinfo{person}{Yu Kang},
  \bibinfo{person}{Xin Gao}, \bibinfo{person}{Liu Shi}, \bibinfo{person}{Yunjie
  Cao}, \bibinfo{person}{Xuedong Gao}, \bibinfo{person}{Hao Fan},
  \bibinfo{person}{Ming Wen}, {et~al\mbox{.}}} \bibinfo{year}{2024}\natexlab{}.
\newblock \showarticletitle{Automatic root cause analysis via large language
  models for cloud incidents}. In \bibinfo{booktitle}{\emph{Proceedings of the
  Nineteenth European Conference on Computer Systems}}.
  \bibinfo{pages}{674--688}.
\newblock


\bibitem[Dragoni et~al\mbox{.}(2017)]%
        {microservices}
\bibfield{author}{\bibinfo{person}{Nicola Dragoni}, \bibinfo{person}{Saverio
  Giallorenzo}, \bibinfo{person}{Alberto~Lluch Lafuente},
  \bibinfo{person}{Manuel Mazzara}, \bibinfo{person}{Fabrizio Montesi},
  \bibinfo{person}{Ruslan Mustafin}, {and} \bibinfo{person}{Larisa Safina}.}
  \bibinfo{year}{2017}\natexlab{}.
\newblock \showarticletitle{Microservices: yesterday, today, and tomorrow}.
\newblock \bibinfo{journal}{\emph{Present and ulterior software engineering}}
  (\bibinfo{year}{2017}), \bibinfo{pages}{195--216}.
\newblock


\bibitem[Edge et~al\mbox{.}(2024)]%
        {graphrag}
\bibfield{author}{\bibinfo{person}{Darren Edge}, \bibinfo{person}{Ha Trinh},
  \bibinfo{person}{Newman Cheng}, \bibinfo{person}{Joshua Bradley},
  \bibinfo{person}{Alex Chao}, \bibinfo{person}{Apurva Mody},
  \bibinfo{person}{Steven Truitt}, \bibinfo{person}{Dasha Metropolitansky},
  \bibinfo{person}{Robert~Osazuwa Ness}, {and} \bibinfo{person}{Jonathan
  Larson}.} \bibinfo{year}{2024}\natexlab{}.
\newblock \showarticletitle{From local to global: A graph rag approach to
  query-focused summarization}.
\newblock \bibinfo{journal}{\emph{arXiv preprint arXiv:2404.16130}}
  (\bibinfo{year}{2024}).
\newblock


\bibitem[Gao et~al\mbox{.}(2023)]%
        {rag_survey}
\bibfield{author}{\bibinfo{person}{Yunfan Gao}, \bibinfo{person}{Yun Xiong},
  \bibinfo{person}{Xinyu Gao}, \bibinfo{person}{Kangxiang Jia},
  \bibinfo{person}{Jinliu Pan}, \bibinfo{person}{Yuxi Bi}, \bibinfo{person}{Yi
  Dai}, \bibinfo{person}{Jiawei Sun}, \bibinfo{person}{Qianyu Guo},
  \bibinfo{person}{Meng Wang}, {et~al\mbox{.}}}
  \bibinfo{year}{2023}\natexlab{}.
\newblock \showarticletitle{Retrieval-Augmented Generation for Large Language
  Models: A Survey}.
\newblock \bibinfo{journal}{\emph{CoRR}} (\bibinfo{year}{2023}).
\newblock


\bibitem[Han et~al\mbox{.}(2024)]%
        {las_rca}
\bibfield{author}{\bibinfo{person}{Yongqi Han}, \bibinfo{person}{Qingfeng Du},
  \bibinfo{person}{Ying Huang}, \bibinfo{person}{Jiaqi Wu},
  \bibinfo{person}{Fulong Tian}, {and} \bibinfo{person}{Cheng He}.}
  \bibinfo{year}{2024}\natexlab{}.
\newblock \showarticletitle{The potential of one-shot failure root cause
  analysis: Collaboration of the large language model and small classifier}. In
  \bibinfo{booktitle}{\emph{Proceedings of the 39th IEEE/ACM International
  Conference on Automated Software Engineering}}. \bibinfo{pages}{931--943}.
\newblock


\bibitem[He et~al\mbox{.}(2017)]%
        {drain}
\bibfield{author}{\bibinfo{person}{Pinjia He}, \bibinfo{person}{Jieming Zhu},
  \bibinfo{person}{Zibin Zheng}, {and} \bibinfo{person}{Michael~R Lyu}.}
  \bibinfo{year}{2017}\natexlab{}.
\newblock \showarticletitle{Drain: An online log parsing approach with fixed
  depth tree}. In \bibinfo{booktitle}{\emph{2017 IEEE international conference
  on web services (ICWS)}}. IEEE, \bibinfo{pages}{33--40}.
\newblock


\bibitem[Hu et~al\mbox{.}(2023)]%
        {gptlens}
\bibfield{author}{\bibinfo{person}{Sihao Hu}, \bibinfo{person}{Tiansheng
  Huang}, \bibinfo{person}{Fatih {\.I}lhan}, \bibinfo{person}{Selim~Furkan
  Tekin}, {and} \bibinfo{person}{Ling Liu}.} \bibinfo{year}{2023}\natexlab{}.
\newblock \showarticletitle{Large language model-powered smart contract
  vulnerability detection: New perspectives}. In \bibinfo{booktitle}{\emph{2023
  5th IEEE International Conference on Trust, Privacy and Security in
  Intelligent Systems and Applications (TPS-ISA)}}. IEEE,
  \bibinfo{pages}{297--306}.
\newblock


\bibitem[Islam et~al\mbox{.}(2024)]%
        {mapcoder}
\bibfield{author}{\bibinfo{person}{Md~Ashraful Islam},
  \bibinfo{person}{Mohammed~Eunus Ali}, {and} \bibinfo{person}{Md~Rizwan
  Parvez}.} \bibinfo{year}{2024}\natexlab{}.
\newblock \showarticletitle{MapCoder: Multi-Agent Code Generation for
  Competitive Problem Solving}. In \bibinfo{booktitle}{\emph{Annual Meeting of
  the Association of Computational Linguistics 2024}}. Association for
  Computational Linguistics (ACL), \bibinfo{pages}{4912--4944}.
\newblock


\bibitem[Jennifer~Mace({[n.\,d.]})]%
        {srebook}
\bibfield{author}{\bibinfo{person}{Stephen Thorne Arup Chakrabarti Jian Ma
  Jessie~Yang Jennifer~Mace, Jelena~Oertel}.}
  \bibinfo{year}{[n.\,d.]}\natexlab{}.
\newblock \bibinfo{title}{SRE Book, Chapter 9: Incident Response}.
\newblock
  \bibinfo{howpublished}{\url{https://sre.google/workbook/incident-response/}}.
\newblock


\bibitem[Jin et~al\mbox{.}(2024)]%
        {mare}
\bibfield{author}{\bibinfo{person}{Dongming Jin}, \bibinfo{person}{Zhi Jin},
  \bibinfo{person}{Xiaohong Chen}, {and} \bibinfo{person}{Chunhui Wang}.}
  \bibinfo{year}{2024}\natexlab{}.
\newblock \showarticletitle{Mare: Multi-agents collaboration framework for
  requirements engineering}.
\newblock \bibinfo{journal}{\emph{arXiv preprint arXiv:2405.03256}}
  (\bibinfo{year}{2024}).
\newblock


\bibitem[Lee et~al\mbox{.}(2023)]%
        {eadro}
\bibfield{author}{\bibinfo{person}{Cheryl Lee}, \bibinfo{person}{Tianyi Yang},
  \bibinfo{person}{Zhuangbin Chen}, \bibinfo{person}{Yuxin Su}, {and}
  \bibinfo{person}{Michael~R Lyu}.} \bibinfo{year}{2023}\natexlab{}.
\newblock \showarticletitle{Eadro: An end-to-end troubleshooting framework for
  microservices on multi-source data}. In \bibinfo{booktitle}{\emph{2023
  IEEE/ACM 45th International Conference on Software Engineering (ICSE)}}.
  IEEE, \bibinfo{pages}{1750--1762}.
\newblock


\bibitem[Lewis et~al\mbox{.}(2020)]%
        {lewis2020retrieval}
\bibfield{author}{\bibinfo{person}{Patrick Lewis}, \bibinfo{person}{Ethan
  Perez}, \bibinfo{person}{Aleksandra Piktus}, \bibinfo{person}{Fabio Petroni},
  \bibinfo{person}{Vladimir Karpukhin}, \bibinfo{person}{Naman Goyal},
  \bibinfo{person}{Heinrich K{\"u}ttler}, \bibinfo{person}{Mike Lewis},
  \bibinfo{person}{Wen-tau Yih}, \bibinfo{person}{Tim Rockt{\"a}schel},
  {et~al\mbox{.}}} \bibinfo{year}{2020}\natexlab{}.
\newblock \showarticletitle{Retrieval-augmented generation for
  knowledge-intensive nlp tasks}.
\newblock \bibinfo{journal}{\emph{Advances in neural information processing
  systems}}  \bibinfo{volume}{33} (\bibinfo{year}{2020}),
  \bibinfo{pages}{9459--9474}.
\newblock


\bibitem[Li et~al\mbox{.}(2022a)]%
        {swisslog}
\bibfield{author}{\bibinfo{person}{Xiaoyun Li}, \bibinfo{person}{Pengfei Chen},
  \bibinfo{person}{Linxiao Jing}, \bibinfo{person}{Zilong He}, {and}
  \bibinfo{person}{Guangba Yu}.} \bibinfo{year}{2022}\natexlab{a}.
\newblock \showarticletitle{SwissLog: Robust anomaly detection and localization
  for interleaved unstructured logs}.
\newblock \bibinfo{journal}{\emph{IEEE Transactions on Dependable and Secure
  Computing}} \bibinfo{volume}{20}, \bibinfo{number}{4} (\bibinfo{year}{2022}),
  \bibinfo{pages}{2762--2780}.
\newblock


\bibitem[Li et~al\mbox{.}(2021)]%
        {tracerca}
\bibfield{author}{\bibinfo{person}{Zeyan Li}, \bibinfo{person}{Junjie Chen},
  \bibinfo{person}{Rui Jiao}, \bibinfo{person}{Nengwen Zhao},
  \bibinfo{person}{Zhijun Wang}, \bibinfo{person}{Shuwei Zhang},
  \bibinfo{person}{Yanjun Wu}, \bibinfo{person}{Long Jiang},
  \bibinfo{person}{Leiqin Yan}, \bibinfo{person}{Zikai Wang}, {et~al\mbox{.}}}
  \bibinfo{year}{2021}\natexlab{}.
\newblock \showarticletitle{Practical root cause localization for microservice
  systems via trace analysis}. In \bibinfo{booktitle}{\emph{2021 IEEE/ACM 29th
  International Symposium on Quality of Service (IWQOS)}}. IEEE,
  \bibinfo{pages}{1--10}.
\newblock


\bibitem[Li et~al\mbox{.}(2022b)]%
        {dejavu}
\bibfield{author}{\bibinfo{person}{Zeyan Li}, \bibinfo{person}{Nengwen Zhao},
  \bibinfo{person}{Mingjie Li}, \bibinfo{person}{Xianglin Lu},
  \bibinfo{person}{Lixin Wang}, \bibinfo{person}{Dongdong Chang},
  \bibinfo{person}{Xiaohui Nie}, \bibinfo{person}{Li Cao},
  \bibinfo{person}{Wenchi Zhang}, \bibinfo{person}{Kaixin Sui},
  {et~al\mbox{.}}} \bibinfo{year}{2022}\natexlab{b}.
\newblock \showarticletitle{Actionable and interpretable fault localization for
  recurring failures in online service systems}. In
  \bibinfo{booktitle}{\emph{Proceedings of the 30th ACM Joint European Software
  Engineering Conference and Symposium on the Foundations of Software
  Engineering}}. \bibinfo{pages}{996--1008}.
\newblock


\bibitem[Liu et~al\mbox{.}(2023)]%
        {g-eval}
\bibfield{author}{\bibinfo{person}{Yang Liu}, \bibinfo{person}{Dan Iter},
  \bibinfo{person}{Yichong Xu}, \bibinfo{person}{Shuohang Wang},
  \bibinfo{person}{Ruochen Xu}, {and} \bibinfo{person}{Chenguang Zhu}.}
  \bibinfo{year}{2023}\natexlab{}.
\newblock \showarticletitle{G-Eval: NLG Evaluation using Gpt-4 with Better
  Human Alignment}. In \bibinfo{booktitle}{\emph{Proceedings of the 2023
  Conference on Empirical Methods in Natural Language Processing}}.
  \bibinfo{pages}{2511--2522}.
\newblock


\bibitem[Ma et~al\mbox{.}(2019)]%
        {ms-rank}
\bibfield{author}{\bibinfo{person}{Meng Ma}, \bibinfo{person}{Weilan Lin},
  \bibinfo{person}{Disheng Pan}, {and} \bibinfo{person}{Ping Wang}.}
  \bibinfo{year}{2019}\natexlab{}.
\newblock \showarticletitle{Ms-rank: Multi-metric and self-adaptive root cause
  diagnosis for microservice applications}. In \bibinfo{booktitle}{\emph{2019
  IEEE International Conference on Web Services (ICWS)}}. IEEE,
  \bibinfo{pages}{60--67}.
\newblock


\bibitem[Ma et~al\mbox{.}(2020)]%
        {automap}
\bibfield{author}{\bibinfo{person}{Meng Ma}, \bibinfo{person}{Jingmin Xu},
  \bibinfo{person}{Yuan Wang}, \bibinfo{person}{Pengfei Chen},
  \bibinfo{person}{Zonghua Zhang}, {and} \bibinfo{person}{Ping Wang}.}
  \bibinfo{year}{2020}\natexlab{}.
\newblock \showarticletitle{Automap: Diagnose your microservice-based web
  applications automatically}. In \bibinfo{booktitle}{\emph{Proceedings of The
  Web Conference 2020}}. \bibinfo{pages}{246--258}.
\newblock


\bibitem[Meng et~al\mbox{.}(2020)]%
        {microcause}
\bibfield{author}{\bibinfo{person}{Yuan Meng}, \bibinfo{person}{Shenglin
  Zhang}, \bibinfo{person}{Yongqian Sun}, \bibinfo{person}{Ruru Zhang},
  \bibinfo{person}{Zhilong Hu}, \bibinfo{person}{Yiyin Zhang},
  \bibinfo{person}{Chenyang Jia}, \bibinfo{person}{Zhaogang Wang}, {and}
  \bibinfo{person}{Dan Pei}.} \bibinfo{year}{2020}\natexlab{}.
\newblock \showarticletitle{Localizing failure root causes in a microservice
  through causality inference}. In \bibinfo{booktitle}{\emph{2020 IEEE/ACM 28th
  International Symposium on Quality of Service (IWQoS)}}. IEEE,
  \bibinfo{pages}{1--10}.
\newblock


\bibitem[Pei et~al\mbox{.}(2025)]%
        {flowofaction}
\bibfield{author}{\bibinfo{person}{Changhua Pei}, \bibinfo{person}{Zexin Wang},
  \bibinfo{person}{Fengrui Liu}, \bibinfo{person}{Zeyan Li},
  \bibinfo{person}{Yang Liu}, \bibinfo{person}{Xiao He}, \bibinfo{person}{Rong
  Kang}, \bibinfo{person}{Tieying Zhang}, \bibinfo{person}{Jianjun Chen},
  \bibinfo{person}{Jianhui Li}, {et~al\mbox{.}}}
  \bibinfo{year}{2025}\natexlab{}.
\newblock \showarticletitle{Flow-of-Action: SOP Enhanced LLM-Based Multi-Agent
  System for Root Cause Analysis}. In \bibinfo{booktitle}{\emph{Companion
  Proceedings of the ACM on Web Conference 2025}}. \bibinfo{pages}{422--431}.
\newblock


\bibitem[Qian et~al\mbox{.}(2025)]%
        {memorag}
\bibfield{author}{\bibinfo{person}{Hongjin Qian}, \bibinfo{person}{Zheng Liu},
  \bibinfo{person}{Peitian Zhang}, \bibinfo{person}{Kelong Mao},
  \bibinfo{person}{Defu Lian}, \bibinfo{person}{Zhicheng Dou}, {and}
  \bibinfo{person}{Tiejun Huang}.} \bibinfo{year}{2025}\natexlab{}.
\newblock \showarticletitle{Memorag: Boosting long context processing with
  global memory-enhanced retrieval augmentation}. In
  \bibinfo{booktitle}{\emph{Proceedings of the ACM on Web Conference 2025}}.
  \bibinfo{pages}{2366--2377}.
\newblock


\bibitem[Reimers and Gurevych(2019)]%
        {sentence-bert}
\bibfield{author}{\bibinfo{person}{Nils Reimers} {and} \bibinfo{person}{Iryna
  Gurevych}.} \bibinfo{year}{2019}\natexlab{}.
\newblock \showarticletitle{Sentence-BERT: Sentence Embeddings using Siamese
  BERT-Networks}. In \bibinfo{booktitle}{\emph{Proceedings of the 2019
  Conference on Empirical Methods in Natural Language Processing}}.
  \bibinfo{publisher}{Association for Computational Linguistics}.
\newblock
\urldef\tempurl%
\url{http://arxiv.org/abs/1908.10084}
\showURL{%
\tempurl}


\bibitem[Roy et~al\mbox{.}(2024)]%
        {react}
\bibfield{author}{\bibinfo{person}{Devjeet Roy}, \bibinfo{person}{Xuchao
  Zhang}, \bibinfo{person}{Rashi Bhave}, \bibinfo{person}{Chetan Bansal},
  \bibinfo{person}{Pedro Las-Casas}, \bibinfo{person}{Rodrigo Fonseca}, {and}
  \bibinfo{person}{Saravan Rajmohan}.} \bibinfo{year}{2024}\natexlab{}.
\newblock \showarticletitle{Exploring llm-based agents for root cause
  analysis}. In \bibinfo{booktitle}{\emph{Companion proceedings of the 32nd ACM
  international conference on the foundations of software engineering}}.
  \bibinfo{pages}{208--219}.
\newblock


\bibitem[Salton and Buckley(1988)]%
        {tfidf}
\bibfield{author}{\bibinfo{person}{Gerard Salton} {and}
  \bibinfo{person}{Christopher Buckley}.} \bibinfo{year}{1988}\natexlab{}.
\newblock \showarticletitle{Term-weighting approaches in automatic text
  retrieval}.
\newblock \bibinfo{journal}{\emph{Information processing \& management}}
  \bibinfo{volume}{24}, \bibinfo{number}{5} (\bibinfo{year}{1988}),
  \bibinfo{pages}{513--523}.
\newblock


\bibitem[Sarmah et~al\mbox{.}(2024)]%
        {hybridrag}
\bibfield{author}{\bibinfo{person}{Bhaskarjit Sarmah}, \bibinfo{person}{Dhagash
  Mehta}, \bibinfo{person}{Benika Hall}, \bibinfo{person}{Rohan Rao},
  \bibinfo{person}{Sunil Patel}, {and} \bibinfo{person}{Stefano Pasquali}.}
  \bibinfo{year}{2024}\natexlab{}.
\newblock \showarticletitle{Hybridrag: Integrating knowledge graphs and vector
  retrieval augmented generation for efficient information extraction}. In
  \bibinfo{booktitle}{\emph{Proceedings of the 5th ACM International Conference
  on AI in Finance}}. \bibinfo{pages}{608--616}.
\newblock


\bibitem[Schulman et~al\mbox{.}(2017)]%
        {ppo}
\bibfield{author}{\bibinfo{person}{John Schulman}, \bibinfo{person}{Filip
  Wolski}, \bibinfo{person}{Prafulla Dhariwal}, \bibinfo{person}{Alec Radford},
  {and} \bibinfo{person}{Oleg Klimov}.} \bibinfo{year}{2017}\natexlab{}.
\newblock \showarticletitle{Proximal policy optimization algorithms}.
\newblock \bibinfo{journal}{\emph{arXiv preprint arXiv:1707.06347}}
  (\bibinfo{year}{2017}).
\newblock


\bibitem[Shinn et~al\mbox{.}(2023)]%
        {reflexion}
\bibfield{author}{\bibinfo{person}{Noah Shinn}, \bibinfo{person}{Federico
  Cassano}, \bibinfo{person}{Ashwin Gopinath}, \bibinfo{person}{Karthik
  Narasimhan}, {and} \bibinfo{person}{Shunyu Yao}.}
  \bibinfo{year}{2023}\natexlab{}.
\newblock \showarticletitle{Reflexion: Language agents with verbal
  reinforcement learning}.
\newblock \bibinfo{journal}{\emph{Advances in Neural Information Processing
  Systems}}  \bibinfo{volume}{36} (\bibinfo{year}{2023}),
  \bibinfo{pages}{8634--8652}.
\newblock


\bibitem[Sui et~al\mbox{.}(2023)]%
        {logkg}
\bibfield{author}{\bibinfo{person}{Yicheng Sui}, \bibinfo{person}{Yuzhe Zhang},
  \bibinfo{person}{Jianjun Sun}, \bibinfo{person}{Ting Xu},
  \bibinfo{person}{Shenglin Zhang}, \bibinfo{person}{Zhengdan Li},
  \bibinfo{person}{Yongqian Sun}, \bibinfo{person}{Fangrui Guo},
  \bibinfo{person}{Junyu Shen}, \bibinfo{person}{Yuzhi Zhang}, {et~al\mbox{.}}}
  \bibinfo{year}{2023}\natexlab{}.
\newblock \showarticletitle{Logkg: Log failure diagnosis through knowledge
  graph}.
\newblock \bibinfo{journal}{\emph{IEEE Transactions on Services Computing}}
  \bibinfo{volume}{16}, \bibinfo{number}{5} (\bibinfo{year}{2023}),
  \bibinfo{pages}{3493--3507}.
\newblock


\bibitem[Sun et~al\mbox{.}(2025a)]%
        {deephunt}
\bibfield{author}{\bibinfo{person}{Yongqian Sun}, \bibinfo{person}{Zihan Lin},
  \bibinfo{person}{Binpeng Shi}, \bibinfo{person}{Shenglin Zhang},
  \bibinfo{person}{Shiyu Ma}, \bibinfo{person}{Pengxiang Jin},
  \bibinfo{person}{Zhenyu Zhong}, \bibinfo{person}{Lemeng Pan},
  \bibinfo{person}{Yicheng Guo}, {and} \bibinfo{person}{Dan Pei}.}
  \bibinfo{year}{2025}\natexlab{a}.
\newblock \showarticletitle{Interpretable failure localization for microservice
  systems based on graph autoencoder}.
\newblock \bibinfo{journal}{\emph{ACM Transactions on Software Engineering and
  Methodology}} \bibinfo{volume}{34}, \bibinfo{number}{2}
  (\bibinfo{year}{2025}), \bibinfo{pages}{1--28}.
\newblock


\bibitem[Sun et~al\mbox{.}(2025b)]%
        {trioxpert}
\bibfield{author}{\bibinfo{person}{Yongqian Sun}, \bibinfo{person}{Yu Luo},
  \bibinfo{person}{Xidao Wen}, \bibinfo{person}{Yuan Yuan},
  \bibinfo{person}{Xiaohui Nie}, \bibinfo{person}{Shenglin Zhang},
  \bibinfo{person}{Tong Liu}, {and} \bibinfo{person}{Xi Luo}.}
  \bibinfo{year}{2025}\natexlab{b}.
\newblock \showarticletitle{TrioXpert: An automated incident management
  framework for microservice system}.
\newblock \bibinfo{journal}{\emph{arXiv preprint arXiv:2506.10043}}
  (\bibinfo{year}{2025}).
\newblock


\bibitem[Sun et~al\mbox{.}(2024)]%
        {art}
\bibfield{author}{\bibinfo{person}{Yongqian Sun}, \bibinfo{person}{Binpeng
  Shi}, \bibinfo{person}{Mingyu Mao}, \bibinfo{person}{Minghua Ma},
  \bibinfo{person}{Sibo Xia}, \bibinfo{person}{Shenglin Zhang}, {and}
  \bibinfo{person}{Dan Pei}.} \bibinfo{year}{2024}\natexlab{}.
\newblock \showarticletitle{Art: A unified unsupervised framework for incident
  management in microservice systems}. In \bibinfo{booktitle}{\emph{Proceedings
  of the 39th IEEE/ACM International Conference on Automated Software
  Engineering}}. \bibinfo{pages}{1183--1194}.
\newblock


\bibitem[Taeb et~al\mbox{.}(2024)]%
        {axnav}
\bibfield{author}{\bibinfo{person}{Maryam Taeb}, \bibinfo{person}{Amanda
  Swearngin}, \bibinfo{person}{Eldon Schoop}, \bibinfo{person}{Ruijia Cheng},
  \bibinfo{person}{Yue Jiang}, {and} \bibinfo{person}{Jeffrey Nichols}.}
  \bibinfo{year}{2024}\natexlab{}.
\newblock \showarticletitle{Axnav: Replaying accessibility tests from natural
  language}. In \bibinfo{booktitle}{\emph{Proceedings of the 2024 CHI
  Conference on Human Factors in Computing Systems}}. \bibinfo{pages}{1--16}.
\newblock


\bibitem[Tao et~al\mbox{.}(2024)]%
        {medicine}
\bibfield{author}{\bibinfo{person}{Lei Tao}, \bibinfo{person}{Shenglin Zhang},
  \bibinfo{person}{Zedong Jia}, \bibinfo{person}{Jinrui Sun},
  \bibinfo{person}{Minghua Ma}, \bibinfo{person}{Zhengdan Li},
  \bibinfo{person}{Yongqian Sun}, \bibinfo{person}{Canqun Yang},
  \bibinfo{person}{Yuzhi Zhang}, {and} \bibinfo{person}{Dan Pei}.}
  \bibinfo{year}{2024}\natexlab{}.
\newblock \showarticletitle{Giving every modality a voice in microservice
  failure diagnosis via multimodal adaptive optimization}. In
  \bibinfo{booktitle}{\emph{Proceedings of the 39th IEEE/ACM International
  Conference on Automated Software Engineering}}. \bibinfo{pages}{1107--1119}.
\newblock


\bibitem[Wang et~al\mbox{.}(2023)]%
        {coral}
\bibfield{author}{\bibinfo{person}{Dongjie Wang}, \bibinfo{person}{Zhengzhang
  Chen}, \bibinfo{person}{Yanjie Fu}, \bibinfo{person}{Yanchi Liu}, {and}
  \bibinfo{person}{Haifeng Chen}.} \bibinfo{year}{2023}\natexlab{}.
\newblock \showarticletitle{Incremental causal graph learning for online root
  cause analysis}. In \bibinfo{booktitle}{\emph{Proceedings of the 29th ACM
  SIGKDD conference on knowledge discovery and data mining}}.
  \bibinfo{pages}{2269--2278}.
\newblock


\bibitem[Wang et~al\mbox{.}(2018)]%
        {cloudranger}
\bibfield{author}{\bibinfo{person}{Ping Wang}, \bibinfo{person}{Jingmin Xu},
  \bibinfo{person}{Meng Ma}, \bibinfo{person}{Weilan Lin},
  \bibinfo{person}{Disheng Pan}, \bibinfo{person}{Yuan Wang}, {and}
  \bibinfo{person}{Pengfei Chen}.} \bibinfo{year}{2018}\natexlab{}.
\newblock \showarticletitle{Cloudranger: Root cause identification for cloud
  native systems}. In \bibinfo{booktitle}{\emph{2018 18th IEEE/ACM
  International Symposium on Cluster, Cloud and Grid Computing (CCGRID)}}.
  IEEE, \bibinfo{pages}{492--502}.
\newblock


\bibitem[Wang et~al\mbox{.}(2024)]%
        {comet}
\bibfield{author}{\bibinfo{person}{Zexin Wang}, \bibinfo{person}{Jianhui Li},
  \bibinfo{person}{Minghua Ma}, \bibinfo{person}{Ze Li}, \bibinfo{person}{Yu
  Kang}, \bibinfo{person}{Chaoyun Zhang}, \bibinfo{person}{Chetan Bansal},
  \bibinfo{person}{Murali Chintalapati}, \bibinfo{person}{Saravan Rajmohan},
  \bibinfo{person}{Qingwei Lin}, {et~al\mbox{.}}}
  \bibinfo{year}{2024}\natexlab{}.
\newblock \showarticletitle{Large language models can provide accurate and
  interpretable incident triage}. In \bibinfo{booktitle}{\emph{2024 IEEE 35th
  International Symposium on Software Reliability Engineering (ISSRE)}}. IEEE,
  \bibinfo{pages}{523--534}.
\newblock


\bibitem[Wei et~al\mbox{.}(2022)]%
        {cot}
\bibfield{author}{\bibinfo{person}{Jason Wei}, \bibinfo{person}{Xuezhi Wang},
  \bibinfo{person}{Dale Schuurmans}, \bibinfo{person}{Maarten Bosma},
  \bibinfo{person}{Fei Xia}, \bibinfo{person}{Ed Chi}, \bibinfo{person}{Quoc~V
  Le}, \bibinfo{person}{Denny Zhou}, {et~al\mbox{.}}}
  \bibinfo{year}{2022}\natexlab{}.
\newblock \showarticletitle{Chain-of-thought prompting elicits reasoning in
  large language models}.
\newblock \bibinfo{journal}{\emph{Advances in neural information processing
  systems}}  \bibinfo{volume}{35} (\bibinfo{year}{2022}),
  \bibinfo{pages}{24824--24837}.
\newblock


\bibitem[Wu et~al\mbox{.}(2021)]%
        {patternmatcher}
\bibfield{author}{\bibinfo{person}{Canhua Wu}, \bibinfo{person}{Nengwen Zhao},
  \bibinfo{person}{Lixin Wang}, \bibinfo{person}{Xiaoqin Yang},
  \bibinfo{person}{Shining Li}, \bibinfo{person}{Ming Zhang},
  \bibinfo{person}{Xing Jin}, \bibinfo{person}{Xidao Wen},
  \bibinfo{person}{Xiaohui Nie}, \bibinfo{person}{Wenchi Zhang},
  {et~al\mbox{.}}} \bibinfo{year}{2021}\natexlab{}.
\newblock \showarticletitle{Identifying root-cause metrics for incident
  diagnosis in online service systems}. In \bibinfo{booktitle}{\emph{2021 IEEE
  32nd International Symposium on Software Reliability Engineering (ISSRE)}}.
  IEEE, \bibinfo{pages}{91--102}.
\newblock


\bibitem[Xie et~al\mbox{.}(2021)]%
        {logm}
\bibfield{author}{\bibinfo{person}{Yuxia Xie}, \bibinfo{person}{Kai Yang},
  {and} \bibinfo{person}{Pan Luo}.} \bibinfo{year}{2021}\natexlab{}.
\newblock \showarticletitle{Logm: Log analysis for multiple components of
  hadoop platform}.
\newblock \bibinfo{journal}{\emph{IEEE Access}}  \bibinfo{volume}{9}
  (\bibinfo{year}{2021}), \bibinfo{pages}{73522--73532}.
\newblock


\bibitem[Xu et~al\mbox{.}(2025)]%
        {openrca}
\bibfield{author}{\bibinfo{person}{Junjielong Xu}, \bibinfo{person}{Qinan
  Zhang}, \bibinfo{person}{Zhiqing Zhong}, \bibinfo{person}{Shilin He},
  \bibinfo{person}{Chaoyun Zhang}, \bibinfo{person}{Qingwei Lin},
  \bibinfo{person}{Dan Pei}, \bibinfo{person}{Pinjia He},
  \bibinfo{person}{Dongmei Zhang}, {and} \bibinfo{person}{Qi Zhang}.}
  \bibinfo{year}{2025}\natexlab{}.
\newblock \showarticletitle{OpenRCA: Can large language models locate the root
  cause of software failures?}. In \bibinfo{booktitle}{\emph{The Thirteenth
  International Conference on Learning Representations}}.
\newblock


\bibitem[Yang et~al\mbox{.}(2023)]%
        {tracenet}
\bibfield{author}{\bibinfo{person}{Jingjing Yang}, \bibinfo{person}{Yuchun
  Guo}, \bibinfo{person}{Yishuai Chen}, {and} \bibinfo{person}{Yongxiang
  Zhao}.} \bibinfo{year}{2023}\natexlab{}.
\newblock \showarticletitle{TraceNet: Operation aware root cause localization
  of microservice system anomalies}. In \bibinfo{booktitle}{\emph{2023 IEEE
  International Conference on Communications Workshops (ICC Workshops)}}. IEEE,
  \bibinfo{pages}{758--763}.
\newblock


\bibitem[Yao et~al\mbox{.}(2023)]%
        {react_yao}
\bibfield{author}{\bibinfo{person}{Shunyu Yao}, \bibinfo{person}{Jeffrey Zhao},
  \bibinfo{person}{Dian Yu}, \bibinfo{person}{Nan Du}, \bibinfo{person}{Izhak
  Shafran}, \bibinfo{person}{Karthik Narasimhan}, {and} \bibinfo{person}{Yuan
  Cao}.} \bibinfo{year}{2023}\natexlab{}.
\newblock \showarticletitle{React: Synergizing reasoning and acting in language
  models}. In \bibinfo{booktitle}{\emph{International Conference on Learning
  Representations (ICLR)}}.
\newblock


\bibitem[Yu et~al\mbox{.}(2023a)]%
        {nezha}
\bibfield{author}{\bibinfo{person}{Guangba Yu}, \bibinfo{person}{Pengfei Chen},
  \bibinfo{person}{Yufeng Li}, \bibinfo{person}{Hongyang Chen},
  \bibinfo{person}{Xiaoyun Li}, {and} \bibinfo{person}{Zibin Zheng}.}
  \bibinfo{year}{2023}\natexlab{a}.
\newblock \showarticletitle{Nezha: Interpretable fine-grained root causes
  analysis for microservices on multi-modal observability data}. In
  \bibinfo{booktitle}{\emph{Proceedings of the 31st ACM Joint European Software
  Engineering Conference and Symposium on the Foundations of Software
  Engineering}}. \bibinfo{pages}{553--565}.
\newblock


\bibitem[Yu et~al\mbox{.}(2023b)]%
        {tracerank}
\bibfield{author}{\bibinfo{person}{Guangba Yu}, \bibinfo{person}{Zicheng
  Huang}, {and} \bibinfo{person}{Pengfei Chen}.}
  \bibinfo{year}{2023}\natexlab{b}.
\newblock \showarticletitle{TraceRank: Abnormal service localization with
  dis-aggregated end-to-end tracing data in cloud native systems}.
\newblock \bibinfo{journal}{\emph{Journal of Software: Evolution and Process}}
  \bibinfo{volume}{35}, \bibinfo{number}{10} (\bibinfo{year}{2023}),
  \bibinfo{pages}{e2413}.
\newblock


\bibitem[Zan et~al\mbox{.}(2024)]%
        {codes}
\bibfield{author}{\bibinfo{person}{Daoguang Zan}, \bibinfo{person}{Ailun Yu},
  \bibinfo{person}{Wei Liu}, \bibinfo{person}{Dong Chen}, \bibinfo{person}{Bo
  Shen}, \bibinfo{person}{Wei Li}, \bibinfo{person}{Yafen Yao},
  \bibinfo{person}{Yongshun Gong}, \bibinfo{person}{Xiaolin Chen},
  \bibinfo{person}{Bei Guan}, {et~al\mbox{.}}} \bibinfo{year}{2024}\natexlab{}.
\newblock \showarticletitle{CodeS: Natural Language to Code Repository via
  Multi-Layer Sketch}.
\newblock \bibinfo{journal}{\emph{CoRR}} (\bibinfo{year}{2024}).
\newblock


\bibitem[Zhang et~al\mbox{.}(2024a)]%
        {paircoder}
\bibfield{author}{\bibinfo{person}{Huan Zhang}, \bibinfo{person}{Wei Cheng},
  \bibinfo{person}{Yuhan Wu}, {and} \bibinfo{person}{Wei Hu}.}
  \bibinfo{year}{2024}\natexlab{a}.
\newblock \showarticletitle{A pair programming framework for code generation
  via multi-plan exploration and feedback-driven refinement}. In
  \bibinfo{booktitle}{\emph{Proceedings of the 39th IEEE/ACM International
  Conference on Automated Software Engineering}}. \bibinfo{pages}{1319--1331}.
\newblock


\bibitem[Zhang et~al\mbox{.}(2023)]%
        {diagfusion}
\bibfield{author}{\bibinfo{person}{Shenglin Zhang}, \bibinfo{person}{Pengxiang
  Jin}, \bibinfo{person}{Zihan Lin}, \bibinfo{person}{Yongqian Sun},
  \bibinfo{person}{Bicheng Zhang}, \bibinfo{person}{Sibo Xia},
  \bibinfo{person}{Zhengdan Li}, \bibinfo{person}{Zhenyu Zhong},
  \bibinfo{person}{Minghua Ma}, \bibinfo{person}{Wa Jin}, {et~al\mbox{.}}}
  \bibinfo{year}{2023}\natexlab{}.
\newblock \showarticletitle{Robust failure diagnosis of microservice system
  through multimodal data}.
\newblock \bibinfo{journal}{\emph{IEEE Transactions on Services Computing}}
  \bibinfo{volume}{16}, \bibinfo{number}{6} (\bibinfo{year}{2023}),
  \bibinfo{pages}{3851--3864}.
\newblock


\bibitem[Zhang et~al\mbox{.}(2024d)]%
        {rca_survey}
\bibfield{author}{\bibinfo{person}{Shenglin Zhang}, \bibinfo{person}{Sibo Xia},
  \bibinfo{person}{Wenzhao Fan}, \bibinfo{person}{Binpeng Shi},
  \bibinfo{person}{Xiao Xiong}, \bibinfo{person}{Zhenyu Zhong},
  \bibinfo{person}{Minghua Ma}, \bibinfo{person}{Yongqian Sun}, {and}
  \bibinfo{person}{Dan Pei}.} \bibinfo{year}{2024}\natexlab{d}.
\newblock \showarticletitle{Failure diagnosis in microservice systems: A
  comprehensive survey and analysis}.
\newblock \bibinfo{journal}{\emph{ACM Transactions on Software Engineering and
  Methodology}} (\bibinfo{year}{2024}).
\newblock


\bibitem[Zhang et~al\mbox{.}(2024c)]%
        {mabc}
\bibfield{author}{\bibinfo{person}{Wei Zhang}, \bibinfo{person}{Hongcheng Guo},
  \bibinfo{person}{Jian Yang}, \bibinfo{person}{Zhoujin Tian},
  \bibinfo{person}{Yi Zhang}, \bibinfo{person}{Yan Chaoran},
  \bibinfo{person}{Zhoujun Li}, \bibinfo{person}{Tongliang Li},
  \bibinfo{person}{Xu Shi}, \bibinfo{person}{Liangfan Zheng}, {et~al\mbox{.}}}
  \bibinfo{year}{2024}\natexlab{c}.
\newblock \showarticletitle{mABC: Multi-Agent Blockchain-inspired Collaboration
  for Root Cause Analysis in Micro-Services Architecture}. In
  \bibinfo{booktitle}{\emph{Findings of the Association for Computational
  Linguistics: EMNLP 2024}}. \bibinfo{pages}{4017--4033}.
\newblock


\bibitem[Zhang et~al\mbox{.}(2024b)]%
        {icl_rca}
\bibfield{author}{\bibinfo{person}{Xuchao Zhang}, \bibinfo{person}{Supriyo
  Ghosh}, \bibinfo{person}{Chetan Bansal}, \bibinfo{person}{Rujia Wang},
  \bibinfo{person}{Minghua Ma}, \bibinfo{person}{Yu Kang}, {and}
  \bibinfo{person}{Saravan Rajmohan}.} \bibinfo{year}{2024}\natexlab{b}.
\newblock \showarticletitle{Automated root causing of cloud incidents using
  in-context learning with GPT-4}. In \bibinfo{booktitle}{\emph{Companion
  Proceedings of the 32nd ACM International Conference on the Foundations of
  Software Engineering}}. \bibinfo{pages}{266--277}.
\newblock


\bibitem[Zhang et~al\mbox{.}(2021)]%
        {onion}
\bibfield{author}{\bibinfo{person}{Xu Zhang}, \bibinfo{person}{Yong Xu},
  \bibinfo{person}{Si Qin}, \bibinfo{person}{Shilin He}, \bibinfo{person}{Bo
  Qiao}, \bibinfo{person}{Ze Li}, \bibinfo{person}{Hongyu Zhang},
  \bibinfo{person}{Xukun Li}, \bibinfo{person}{Yingnong Dang},
  \bibinfo{person}{Qingwei Lin}, {et~al\mbox{.}}}
  \bibinfo{year}{2021}\natexlab{}.
\newblock \showarticletitle{Onion: identifying incident-indicating logs for
  cloud systems}. In \bibinfo{booktitle}{\emph{Proceedings of the 29th ACM
  Joint Meeting on European Software Engineering Conference and Symposium on
  the Foundations of Software Engineering}}. \bibinfo{pages}{1253--1263}.
\newblock


\bibitem[Zheng et~al\mbox{.}(2024)]%
        {mulan}
\bibfield{author}{\bibinfo{person}{Lecheng Zheng}, \bibinfo{person}{Zhengzhang
  Chen}, \bibinfo{person}{Jingrui He}, {and} \bibinfo{person}{Haifeng Chen}.}
  \bibinfo{year}{2024}\natexlab{}.
\newblock \showarticletitle{MULAN: multi-modal causal structure learning and
  root cause analysis for microservice systems}. In
  \bibinfo{booktitle}{\emph{Proceedings of the ACM Web Conference 2024}}.
  \bibinfo{pages}{4107--4116}.
\newblock


\bibitem[Zhou et~al\mbox{.}(2024)]%
        {dbot}
\bibfield{author}{\bibinfo{person}{Xuanhe Zhou}, \bibinfo{person}{Guoliang Li},
  \bibinfo{person}{Zhaoyan Sun}, \bibinfo{person}{Zhiyuan Liu},
  \bibinfo{person}{Weize Chen}, \bibinfo{person}{Jianming Wu},
  \bibinfo{person}{Jiesi Liu}, \bibinfo{person}{Ruohang Feng}, {and}
  \bibinfo{person}{Guoyang Zeng}.} \bibinfo{year}{2024}\natexlab{}.
\newblock \showarticletitle{D-Bot: Database Diagnosis System using Large
  Language Models}.
\newblock \bibinfo{journal}{\emph{Proceedings of the VLDB Endowment}}
  \bibinfo{volume}{17}, \bibinfo{number}{10} (\bibinfo{year}{2024}),
  \bibinfo{pages}{2514--2527}.
\newblock


\bibitem[Zhou et~al\mbox{.}(2019)]%
        {mepfl}
\bibfield{author}{\bibinfo{person}{Xiang Zhou}, \bibinfo{person}{Xin Peng},
  \bibinfo{person}{Tao Xie}, \bibinfo{person}{Jun Sun}, \bibinfo{person}{Chao
  Ji}, \bibinfo{person}{Dewei Liu}, \bibinfo{person}{Qilin Xiang}, {and}
  \bibinfo{person}{Chuan He}.} \bibinfo{year}{2019}\natexlab{}.
\newblock \showarticletitle{Latent error prediction and fault localization for
  microservice applications by learning from system trace logs}. In
  \bibinfo{booktitle}{\emph{Proceedings of the 2019 27th ACM joint meeting on
  European software engineering conference and symposium on the foundations of
  software engineering}}. \bibinfo{pages}{683--694}.
\newblock


\end{thebibliography}

\end{document}